\documentclass[11pt, a4paper, goog]{google}

\renewcommand{\today}{2026-10-08}

\usepackage[authoryear, sort&compress, round]{natbib}
\usepackage{amsmath,amsfonts,bm}

\def\eqref#1{equation~\ref{#1}}

\def\1{\bm{1}}

\DeclareMathAlphabet{\mathsfit}{\encodingdefault}{\sfdefault}{m}{sl}
\SetMathAlphabet{\mathsfit}{bold}{\encodingdefault}{\sfdefault}{bx}{n}

\usepackage{hyperref}
\usepackage{url}
\usepackage{graphicx}
\usepackage{booktabs}
\usepackage{tabularx}
\usepackage{amsmath,amssymb}
\usepackage{algorithm}
\usepackage{algorithmic}
\usepackage{xcolor}
\usepackage{colortbl}

\definecolor{dtddrow}{RGB}{231,243,250}
\definecolor{teacherrow}{RGB}{253,241,197}
\definecolor{settingrow}{RGB}{244,244,244}
\definecolor{resultgreen}{RGB}{0,120,70}
\definecolor{resultred}{RGB}{190,45,45}

\newcommand{\dtdd}{\textsc{DTDD}}
\newcommand{\opd}{OPD}
\newcommand{\best}[1]{\textbf{#1}}
\newcommand{\posdiff}[1]{\hspace{0.1em}\raisebox{-0.3ex}{\textcolor{resultgreen}{\fontsize{4.5}{5}\selectfont $+#1$}}}
\newcommand{\negdiff}[1]{\hspace{0.1em}\raisebox{-0.3ex}{\textcolor{resultred}{\fontsize{4.5}{5}\selectfont $-#1$}}}

\newif\ificlr


\uselogo{}

\title{DTDD: Divergence-Triggered Dynamic Distillation for Reliable
On-Policy Supervision}

\correspondingauthor{heqj3@uw.edu, sugato@google.com}

\makeatletter
\apptocmd{\ps@firststyle}{%
  \fancyfoot[L]{%
    \footerfont\textsuperscript{*}This work was done while Qijia was an intern at Google Ads AI.\\
    {\footerfont\itshape Corresponding author(s): \the\correspondingauthor}%
  }%
}{}{\PackageError{DTDD}{Could not extend the first-page footer}{}}
\makeatother

\author[1,2,*]{Qijia He}
\author[1]{Zequn Li}
\author[1]{Spencer Luo}
\author[1,3]{Tianle Chen}
\author[1]{Pradyumna Narayana}
\author[2]{Zixian Ma}
\author[2]{Ranjay Krishna}
\author[1]{Lan Nie}
\author[1]{Sugato Basu}

\let\affiloriginalrlap\rlap
\DeclareRobustCommand{\affillogoGoogle}{%
  \begingroup\let\rlap\affiloriginalrlap
  \,\raisebox{-0.18em}{\includegraphics[height=0.95em]{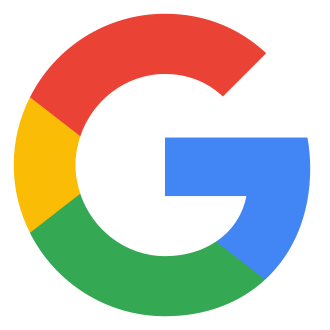}}\,%
  \endgroup}
\DeclareRobustCommand{\affillogoUW}{%
  \begingroup\let\rlap\affiloriginalrlap
  \,\raisebox{-0.18em}{\includegraphics[height=0.95em]{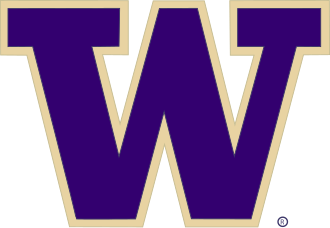}}\,%
  \endgroup}
\DeclareRobustCommand{\affillogoBU}{%
  \begingroup\let\rlap\affiloriginalrlap
  \,\raisebox{-0.18em}{\includegraphics[height=0.95em]{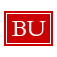}}\,%
  \endgroup}
\affil[1]{\affillogoGoogle Google}
\affil[2]{\affillogoUW University of Washington}
\affil[3]{\affillogoBU Boston University}

\usepackage{fontawesome5}
\newcommand{\projecturl}{https://dtdd-proj.github.io}
\makeatletter
\pretocmd{\abscontent}{%
  \vspace*{-22pt}\noindent{\Affilfont\faGlobe\hspace{0.35em}Project page:
  \href{\projecturl}{dtdd-proj.github.io}}\par\vskip6pt}%
  {}{\PackageError{DTDD}{Could not add the project-page line}{}}
\makeatother

\begin{abstract}
On-policy distillation (OPD) provides dense token-level supervision on student
rollouts, but this supervision can become unreliable when the student and
teacher diverge, either because of a large initial policy gap or compounding
errors in long-horizon reasoning and tool use. We introduce
Divergence-Triggered Dynamic Distillation (\dtdd{}),
an adaptive OPD framework that uses segment-level student--teacher divergence
to trigger teacher takeover during training.
When divergence triggers an intervention during reasoning, the student prefix
continues to receive standard OPD supervision. The teacher then generates a
recovery suffix, which the student learns through behavior cloning, using
cross-entropy on the teacher-generated tokens. 
We evaluate \dtdd{} in three regimes: long-horizon reasoning, distillation from
a base initialization without offline supervised fine-tuning (SFT), and
multi-turn agents. We also
compare alternative trigger designs and teacher-intervention rates. \dtdd{}
reduces the post-warmup peak gradient norm by $8.6\times$ and, from a base
initialization, reaches $21.3\%$ held-out math accuracy in $2.6\times$ fewer
optimizer updates than OPD; both methods finish near $22\%$. It improves
best-observed performance by $4.8$ and $4.1$
percentage points on ALFWorld with LoRA and agentic code, respectively.
Ablations show that segment-averaged triggers outperform token-spike triggers.
Behavior cloning is also more effective than the tested OPD surrogate for training
teacher-generated recoveries. Across 23 experimental arms, takeover tends to
help
when the teacher substantially outperforms the OPD-trained student, provided
that the teacher can recover from student-visited states, the student can learn
those recoveries, and teacher intervention is neither excessive nor
insufficient.
\end{abstract}

\begin{document}

\maketitle
\suppressfloats[t]

\section{Introduction}
\label{sec:intro}

\begin{figure*}[t]
  \centering
\includegraphics[width=0.8\textwidth]{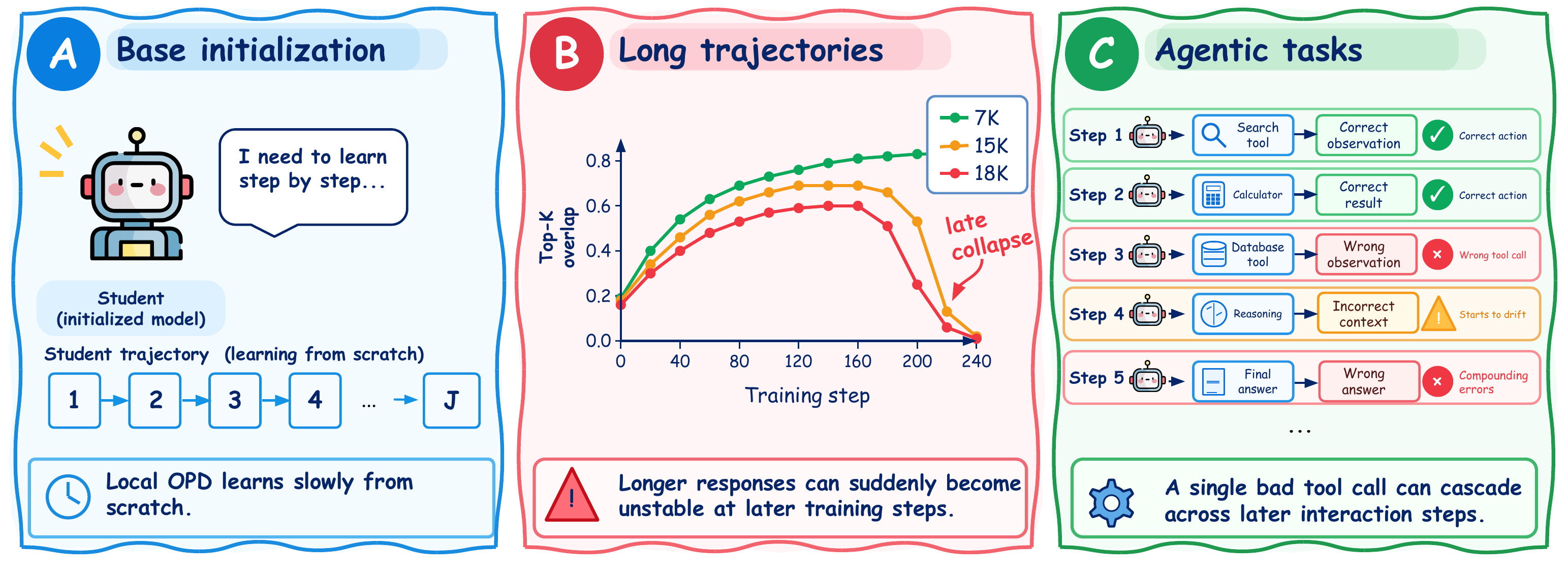}
  \caption{Representative failures of local OPD supervision:
  \textbf{(A)}~slow learning from a base initialization, \textbf{(B)}~late
  instability under long generation budgets, and \textbf{(C)}~compounding errors
  after an incorrect action in multi-turn interaction. Appendix~\ref{app:opd-limitations}
  reports measured examples of (A) and (B).}
  \label{fig:opd-failure-cases}
\end{figure*}

Knowledge distillation transfers capabilities from a strong teacher to a
smaller student \citep{kd}. Training on fixed teacher-generated sequences
creates a distribution mismatch: the student learns from teacher-generated
prefixes during training but must condition on its own prefixes at inference
\citep{seqkd,gkd}. One motivation for on-policy distillation (OPD) is to reduce
this state-distribution mismatch by sampling student trajectories and querying
the teacher at student-visited states \citep{gkd}. OPD therefore combines
student-state coverage with dense teacher supervision. The quality of this
signal depends on whether the teacher provides learnable guidance along the
states the student visits.

Reliable on-policy supervision can fail for two broad reasons: the teacher
signal may be poorly matched to the student at initialization, or it may
deteriorate as errors accumulate along a rollout. In reasoning, incompatible
student--teacher thinking patterns and limited teacher information gain can make
early OPD ineffective, motivating supervised fine-tuning (SFT) on teacher
rollouts and teacher-aligned prompts
\citep{rethinkingopd}. Even when initial learning is viable, longer reasoning
rollouts can amplify deviations and destabilize training \citep{stableopd}. In
multi-turn tool use, an incorrect tool call additionally changes subsequent
observations and contexts, causing errors and student--teacher divergence to
compound across interactions \citep{sod,tcod}.
Figure~\ref{fig:opd-failure-cases} illustrates these cases. Despite their
different origins, each can move the student into prefixes where local
reverse-KL feedback is no longer informative \citep{tgpo,revisitingopd}. OPD
ensures coverage of student-visited states, but effective learning still
requires a reliable and learnable signal at those states.

Prior work addresses unreliable OPD supervision through two broad strategies.
The first modifies the token-level learning signal while retaining the student
trajectory: StableOPD constrains policy drift and mixes in fixed demonstrations,
TGPO introduces directional teacher guidance, and SOD attenuates distillation
in high-divergence regions \citep{stableopd,tgpo,sod}. These methods protect or
redirect optimization within the student trajectory. The second strategy
introduces coherent teacher-generated corrections, including
curriculum-controlled trajectory segments, expert suffixes, rewritten
trajectories, per-step teacher targets, and temporary teacher relays
\citep{tcod,oec,guidedopd,trd,sopd,relayopd,adaswitch}. Together, these approaches
motivate studying where to intervene and how to train on the resulting
trajectory. We investigate segment-averaged divergence as a boundary for
retaining OPD on the student prefix and applying behavior cloning to a
teacher-generated recovery suffix.

To this end, we propose \textbf{DTDD}, a
\textbf{D}ivergence-\textbf{T}riggered \textbf{D}ynamic
\textbf{D}istillation framework built around a simple teacher-takeover rule
during training.
\dtdd{} partitions each student rollout into short token segments and measures
student--teacher divergence within each segment. When a segment exhibits high
divergence, \dtdd{} rewinds to its beginning and lets the teacher complete the
trajectory from the retained student prefix. The prefix continues to receive
standard OPD supervision,
while the teacher-generated suffix is trained with behavior cloning, using
cross-entropy on teacher-generated tokens. Without a trigger, \dtdd{} reduces
to standard OPD. The same
mechanism can intervene sparsely to stabilize an already capable student or
more frequently to provide coherent supervision from a base initialization. In
multi-turn settings, takeover instead replaces the current
reasoning-and-tool-call action before interaction continues.

We evaluate \dtdd{} on long-horizon math, base-model distillation without
offline SFT, and multi-turn agents. With 15K- and 18K-token response budgets,
it preserves held-out accuracy while reducing post-warmup peak gradient norms
by $8.6\times$ and $7.0\times$. From
Qwen3-1.7B-Base, high-dose \dtdd{}
reaches $21.3\%$ held-out accuracy after roughly 70 updates and stays near that
level; OPD reaches it around update 180, and both finish near $22\%$. At
comparable accuracy, \dtdd{} requires 22.0 serial hours and 43M teacher-written
tokens; offline SFT followed by OPD requires 42.8 hours and 95M. Action-level
takeover improves best-observed performance by $4.8$ and $4.1$ percentage
points on ALFWorld with LoRA and agentic code, respectively. Ablations isolate
two design choices: the takeover trigger and the teacher-suffix loss.
Segment-averaged divergence outperforms individual token spikes, and behavior
cloning outperforms the tested OPD surrogate on teacher suffixes. Across 23 arms in 13
settings, 19 follow a common pattern. \dtdd{} tends to help when the teacher is
stronger than the OPD student, can recover from student prefixes, and provides
continuations the student can learn; the intervention rate must provide enough
guidance without replacing on-policy training.

\section{Related Work}
\label{sec:related}

\paragraph{On-policy distillation.}
Knowledge distillation transfers teacher knowledge through output-level
supervision \citep{kd}. Conventional off-policy approaches either match
teacher token distributions under fixed contexts or train on complete
teacher-generated sequences \citep{seqkd}. On-policy distillation instead
computes teacher supervision along student-generated trajectories
\citep{minillm,gkd}. Recent analyses identify thinking-pattern incompatibility,
limited teacher information gain, rollout-length effects, and unreliable
sampled-token feedback as
important failure sources \citep{rethinkingopd,revisitingopd}. To reduce initial
pattern mismatch, \citet{rethinkingopd} first fine-tune the student on offline
teacher rollouts before applying OPD. StableOPD \citep{stableopd}
combines a reference-based divergence constraint with auxiliary SFT on fixed,
high-quality ``golden'' demonstrations to counter repetition-driven length
inflation. Other methods reshape supervision within the student rollout: SOD
\citep{sod} reweights interaction steps using step-level divergence, while
Prune-OPD \citep{pruneopd} downweights and truncates a rollout after detecting
prefix drift. These approaches improve initialization, augment the training
distribution, or modify the original learning signal. \dtdd{} complements them
by conditionally regenerating an unreliable region of the current student
trajectory.

\paragraph{Teacher-generated trajectory intervention.}
A complementary line of work changes the training trajectory itself by
introducing teacher-generated corrections. These methods differ primarily in
how the intervention boundary is chosen. TRD \citep{trd} rewrites complete
student rollouts, whereas SOPD \citep{sopd} generates a teacher target step
from each student-visited step prefix. Other methods allocate trajectory
control independently of the current student--teacher discrepancy. OEC
\citep{oec} uses a random one-way expert handoff; TCOD \citep{tcod} schedules
the horizon controlled by the student, with its backward-to-forward curriculum
replaying prefixes from pre-collected successful teacher trajectories.
More closely related to our setting, Relay-OPD \citep{relayopd} and DASH-OPD
\citep{dashopd} use state-dependent signals to insert temporary teacher
segments. DASH-OPD also uses different distillation objectives for student-
and teacher-generated turns. AdaSwitch \citep{adaswitch} uses a context-relative
next-token divergence spike to trigger a one-way teacher takeover and optimizes
the resulting hybrid sequence with a unified KD objective. Compared with
AdaSwitch, \dtdd{} makes two design changes. First, it averages divergence over
short token segments and triggers at the first segment whose average exceeds
the threshold, rather than reacting to an individual token spike. Second, in
reasoning tasks, \dtdd{} rewinds to that segment, retains the preceding student
prefix, and asks the teacher to regenerate the remaining suffix. The student
prefix is trained with OPD, while the teacher-generated suffix is trained with
behavior cloning. In multi-turn tasks, the same trigger replaces the current
action or transfers the remaining episode to the teacher, depending on the
interactive realization.
At inference time, ARCHER \citep{archer} chooses among reflection, replanning,
and escalation to a stronger model after execution failures. \dtdd{} uses
teacher recovery as supervision during student training.

\section{Divergence-Triggered Dynamic Distillation}
\label{sec:method}

We first formulate the OPD objective, then present \dtdd{} and formalize its
segment-level trigger, teacher takeover, and training objective.

\subsection{On-Policy Distillation Preliminaries}
\label{sec:preliminaries}

Let $x\sim\mathcal D_x$ be a prompt, $\pi_\theta$ the student, and
$\pi_T$ a frozen teacher. The student samples an autoregressive rollout
$y^S=(y^S_1,\ldots,y^S_L)\sim\pi_\theta(\cdot\mid x)$ under a maximum generation
budget $R$, where $L\le R$. At position $t$, let
$h_t=(x,y^S_{<t})$ denote the student-visited history, with
$p_t(v)=\pi_\theta(v\mid h_t)$ and $q_t(v)=\pi_T(v\mid h_t)$ denoting the
student and teacher next-token distributions. On-policy distillation evaluates
the teacher along trajectories sampled from the current student \citep{gkd}.
We use a reverse-KL objective \citep{minillm}, writing
$\ell_t^{\mathrm{RKL}}(\theta)=D_{\mathrm{KL}}(p_t\|q_t)$ for its
per-position loss:
\begin{equation}
  \mathcal L_{\mathrm{OPD}}(\theta)
  = \mathbb E_{\substack{x\sim\mathcal D_x,\\
                         y^S\sim\pi_\theta(\cdot\mid x)}}
    \left[\frac{1}{L}\sum_{t=1}^{L}
    \ell_t^{\mathrm{RKL}}(\theta)\right].
  \label{eq:opd-objective}
\end{equation}
To obtain multi-token supervision without evaluating the full vocabulary, we
retain the student's $K$ highest-probability next-token candidates. Let
$\mathcal V_t=\operatorname{TopK}(p_t,K)$ and
$\bar p_t(v)=p_t(v)/\sum_{u\in\mathcal V_t}p_t(u)$ for
$v\in\mathcal V_t$. For each candidate, we construct the token-level signal
\begin{equation}
  A^{\mathrm{OPD}}_{t,v}
  = \bar p_t(v)\bigl(\log q_t(v)-\log p_t(v)\bigr).
  \label{eq:opd-topk-advantage}
\end{equation}
A positive signal increases the probability of a candidate favored more
strongly by the teacher, whereas a negative signal suppresses a candidate
over-preferred by the student. We treat these signals as detached token-level
advantages during optimization. The corresponding unclipped surrogate is
\begin{equation}
  \widehat\ell_t^{\mathrm{OPD}}(\theta)
  =-\sum_{v\in\mathcal V_t}
    \operatorname{sg}\!\left[A^{\mathrm{OPD}}_{t,v}\right]
    \log \pi_\theta(v\mid h_t),
  \label{eq:opd-surrogate}
\end{equation}
where $\operatorname{sg}$ denotes stop-gradient. This is the policy-gradient
surrogate used to optimize the top-$K$ approximation of reverse KL. We use it
for reasoning OPD, while
Section~\ref{sec:trigger} defines the separate nonnegative discrepancy used for
triggering.

\subsection{DTDD Overview}
\label{sec:motivation}

\begin{figure*}[t]
  \centering
  \includegraphics[width=0.95\textwidth]{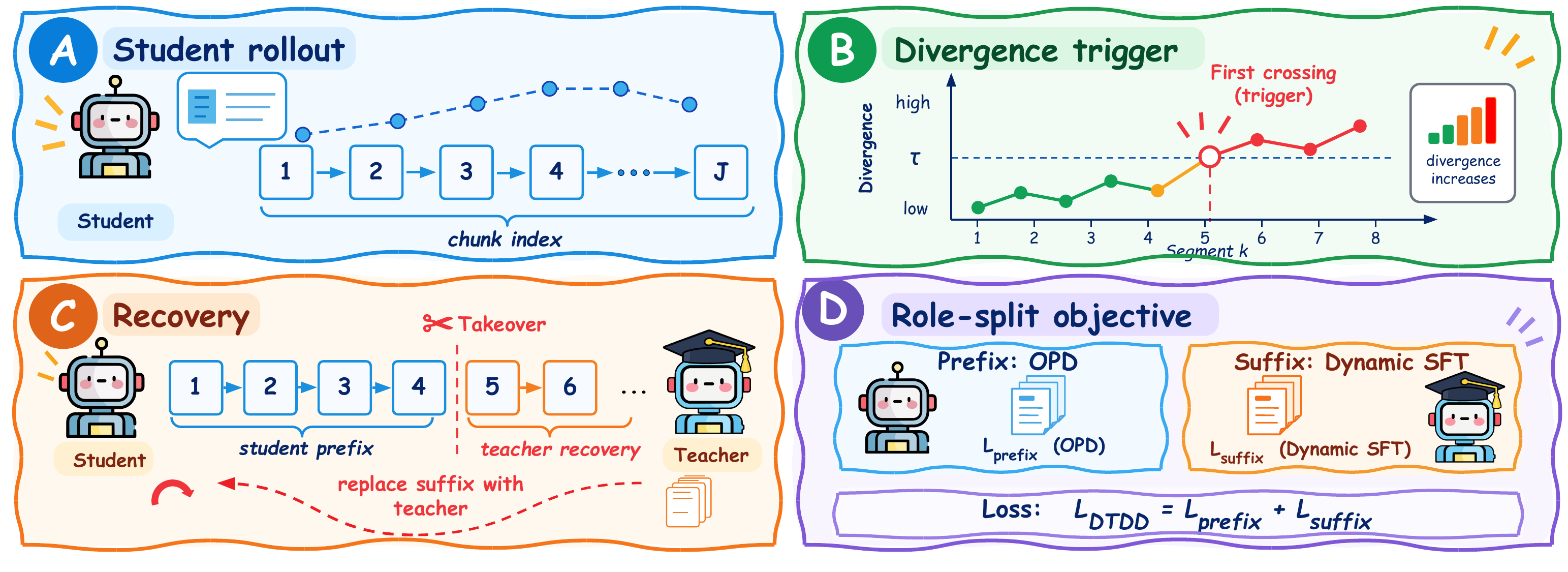}
  \caption{Reasoning realization of \dtdd{}. (A) The student generates a
  segmented rollout. (B) The first threshold crossing selects the rewind
  boundary. (C) The teacher regenerates the suffix from the retained student
  prefix. (D) The prefix is trained with OPD and the suffix with behavior
  cloning; without a crossing, \dtdd{} reduces to OPD.}
  \label{fig:dtdd-overview}
\end{figure*}

\dtdd{} routes supervision in three stages (Figure~\ref{fig:dtdd-overview}).
First, the student generates an on-policy rollout partitioned into fixed-length
segments. A prespecified controller, configured for the desired intervention
regime, selects the first segment whose aggregated student--teacher divergence
crosses its threshold. Second, \dtdd{} rewinds to the preceding boundary and
asks the frozen teacher to regenerate the remaining suffix from the retained
student prefix. Third, the student prefix receives OPD supervision, while the
teacher suffix receives behavior cloning. If no crossing occurs, the entire
rollout receives standard OPD.

The controller setting determines the intervention dose. Frequent takeover
supplies coherent trajectory-level guidance when a base student has not
acquired a teacher-compatible reasoning pattern, whereas sparse takeover
repairs localized failures in otherwise effective OPD. In multi-turn tasks,
the same routing logic operates at action boundaries, where the teacher
generates from the triggering state before interaction continues.

\subsection{Segment-Level Trigger and Recovery}
\label{sec:trigger}

We now define the segment discrepancy, intervention boundary, and recovery
trajectory. Algorithm~\ref{alg:dtdd} summarizes the reasoning realization.

We construct the trigger from a nonnegative token-level discrepancy that
captures the magnitude of local disagreement rather than its direction. In our
reasoning experiments, it is computed from the same student top-$K$
comparisons as the sparse OPD estimator:
\begin{equation}
  \delta_t
  = \sum_{v\in\mathcal V_t}\bar p_t(v)
    \left|\log p_t(v)-\log q_t(v)\right|.
  \label{eq:token-divergence}
\end{equation}
Let $(W_1,\ldots,W_J)$ be an ordered partition of the
model-generated positions into \emph{segments}. We define the discrepancy of
segment $j$ as
\begin{equation}
  d_j = \frac{1}{|W_j|}\sum_{t\in W_j}\delta_t.
  \label{eq:segment-divergence}
\end{equation}
For a reasoning rollout, we use fixed-length segments of $C$ tokens:
$J=\lceil L/C\rceil$ and
$W_j=\{(j-1)C+1,\ldots,\min(jC,L)\}$. For an agentic trajectory, each $W_j$
instead contains the tokens of the $j$-th model-generated action; environment
observations provide context but are not included in the discrepancy.

Let $j_{\min}$ be the first eligible segment and $\tau_s$ the threshold at
optimization step $s$. \dtdd{} selects the first eligible threshold crossing,
$
  j^\star
  = \min\left\{j\ge j_{\min}:d_j>\tau_s\right\}$.
If no eligible segment crosses the threshold, \dtdd{} leaves the trajectory
unchanged and applies standard OPD. Averaging divergence within a segment
prevents isolated token-level spikes from causing intervention.

We instantiate $\tau_s$ with either a fixed or rate-controlled threshold. Both
variants use divergence to select which rollouts trigger and the first crossing
to determine the intervention boundary. Appendix~\ref{app:controller-details}
gives the controller and recovery details.

In reasoning tasks, when a trigger occurs, we define the rewind
position $c=(j^\star-1)C$ as the boundary immediately preceding the triggering
segment.
\dtdd{} then replaces the student suffix $y^S_{c+1:L}$ with a continuation
$y^T_{c+1:\widetilde L}$ generated by the frozen teacher conditioned on
$(x,y^S_{1:c})$ within the remaining generation budget. The resulting hybrid
trajectory is
$
  \widetilde y
  = \bigl[y^S_{1:c};\,y^T_{c+1:\widetilde L}\bigr].$
Thus the teacher replaces the divergent segment itself, rather than inheriting
it and correcting only later tokens. Once control is transferred, the teacher
completes the suffix without returning generation to the student. If no trigger
occurs, we set $c=L$ and $\widetilde y=y^S$, so \dtdd{} reduces exactly to OPD.

The recovery horizon follows how errors propagate. In agentic code, later tool
calls depend on earlier outputs, so errors can compound. When a student action
triggers intervention, that action is discarded and the teacher takes over
from that point, generating the replacement action and the rest of the episode.
In ALFWorld and WebShop, the teacher replaces only the triggering action, and
the student resumes at the next turn from the resulting environment state.
We use this shorter recovery horizon because each executed action yields a new
observation that provides a natural point for returning control to the student.
Teacher-written tokens receive behavior cloning in agentic code, as in
reasoning. ALFWorld and WebShop instead reuse their sampled-token OPD
objective for both student- and teacher-written actions, weighting each token
by its teacher--student log-probability gap.
Appendix~\ref{app:interactive-protocol} gives the corresponding objectives.

\begin{algorithm}[!t]
  \caption{\dtdd{} Training for Reasoning Rollouts}
  \label{alg:dtdd}
  \begin{algorithmic}[1]
    \REQUIRE Student $\pi_\theta$; frozen teacher $\pi_T$; threshold
    controller $H$; rollout parameters $R,C,j_{\min}$
    \FOR{$s=1,\ldots,S$}
      \STATE Roll out $\mathcal B=\{(x_i,y_i^S)\}$ with
      $y_i^S\sim\pi_\theta(\cdot\mid x_i)$
      \STATE Compute $d_{i,1:J_i}$ by
      Equations~\ref{eq:token-divergence}--\ref{eq:segment-divergence};
      $\tau_s\leftarrow H_s(\{d_{i,j}\}_{i,j})$
      \FOR{each $(x_i,y_i^S)\in\mathcal B$}
        \STATE $j_i^\star\leftarrow$ first $j\ge j_{\min}$ with
        $d_{i,j}>\tau_s$, if one exists
        \IF{no such $j_i^\star$ exists}
          \STATE $(c_i,\widetilde y_i)\leftarrow(L_i,y_i^S)$
          \COMMENT{standard OPD}
        \ELSE
          \STATE $c_i\leftarrow(j_i^\star-1)C$
          \COMMENT{rewind}
          \STATE Sample $y_i^T\sim\pi_T(\cdot\mid x_i,y^S_{i,1:c_i})$;
          $\widetilde y_i\leftarrow[y^S_{i,1:c_i};y_i^T]$
          \COMMENT{teacher recovery}
        \ENDIF
      \ENDFOR
      \STATE Update $\theta$ using $\mathcal L_{\mathrm{DTDD}}$:
      OPD on student prefixes and behavior cloning on teacher suffixes
    \ENDFOR
  \end{algorithmic}
\end{algorithm}

\subsection{Role-Split Training Objective}
\label{sec:objective}

For a recovered rollout, \dtdd{} applies OPD to the retained student prefix
and behavior cloning to the teacher-generated suffix. Behavior cloning is
standard token-level cross-entropy: each teacher-written token is the target,
conditioned on the preceding hybrid trajectory. Let
$\widehat\ell_t^{\mathrm{OPD}}(\theta)$ be the sparse OPD surrogate from
Equation~\ref{eq:opd-surrogate}. The suffix loss is
$\ell_t^{\mathrm{BC}}(\theta)=-\log
\pi_\theta(\widetilde y_t\mid x,\widetilde y_{<t})$, and the combined objective is
\begin{equation}
  \mathcal L_{\mathrm{DTDD}}(\theta)
  = \mathbb E\left[\frac{1}{\widetilde L}\left(
      \sum_{t=1}^{c}\widehat\ell_t^{\mathrm{OPD}}(\theta)
      + \lambda_{\mathrm{BC}}
        \sum_{t=c+1}^{\widetilde L}\ell_t^{\mathrm{BC}}(\theta)
    \right)\right],
  \label{eq:dtdd-objective}
\end{equation}
where $\lambda_{\mathrm{BC}}$ weights suffix supervision. If no
trigger occurs, $c=L$ and $\widetilde y=y^S$, so the objective reduces to OPD.
We use $\lambda_{\mathrm{BC}}=1$ in the
main experiments. Section~\ref{sec:setup}
(Table~\ref{tab:suffix-objective}) varies this weight from $0.1$ to $3$ and
also evaluates a soft cross-entropy target.

\section{Experiments}
\label{sec:experiments}

We evaluate \dtdd{} across long-form reasoning and multi-turn agents, asking
whether takeover improves long-horizon stability (\textbf{Q1}), base-model
learning efficiency (\textbf{Q2}), transfer to interactive tasks (\textbf{Q3}),
and which student--teacher regimes support intervention (\textbf{Q4}).

\subsection{Experimental Setup}
\label{sec:setup}

\paragraph{Evaluation settings.}
We evaluate \dtdd{} in two complementary regimes. The reasoning suite tests
long-horizon stability by distilling DeepSeek-R1-Distill-Qwen-1.5B from
JustRL-DeepSeek-1.5B on DAPO-Math-17K
\citep{deepseekr1,dapo}, and
base-initialization efficiency by pairing Qwen3-1.7B-Base with
Qwen3-4B-Base-GRPO on DAPO-Math-17K and with a Qwen3-4B non-thinking teacher
on OpenThoughts3 \citep{qwen3,openthoughts}. Both math settings follow the
experimental configurations in \citet{rethinkingopd}. The interactive suite
tests action-level takeover on ALFWorld \citep{alfworld} with a
Qwen2.5-3B-Instruct student \citep{qwen25} and GiGPO-7B teacher
\citep{gigpo}, and on the code subset of
Open-AgentRL-30K with a Qwen3-1.7B-SFT student and DemyAgent-GRPO-4B teacher.

\paragraph{Comparison and metrics.}
Vanilla OPD is the direct baseline. We also evaluate AdaSwitch
\citep{adaswitch} in the base-student setting
(Table~\ref{tab:main-base-controls}; Appendix~\ref{app:base-controls}).
Within each setting, paired OPD and \dtdd{} runs share the
student initialization, teacher, prompts, batch order, generation budget, and
optimizer, providing a controlled comparison of the intervention controller. For
training-dynamics and ablation claims, we compare fixed shared updates or the
full learning curves. Table~\ref{tab:main-results} reports the best observed
checkpoint for each method, selected by Average within its block. Appendix
Table~\ref{tab:app-matched-endpoint} reports shared-update comparisons;
Appendices~\ref{app:stability-dynamics} and~\ref{app:base-dose-response}
summarize the corresponding reasoning curves.
We report
average@16 accuracy on AIME 2024, AIME 2025, and AMC 2023 for math; success on
the seen and unseen splits for ALFWorld; and average@8 execution accuracy on
LiveCodeBench \citep{livecodebench} and held-out in-distribution problems for
agentic code. Appendix~\ref{app:experimental-details} specifies the full
training, sampling, checkpoint-selection, and reporting protocols.

\paragraph{Design controls.}
Table~\ref{tab:main-base-controls} ablates two choices in the base-student
setting: the triggering mechanism and the teacher-suffix loss. We cross
segment-averaged divergence and token-spike triggers with BC (cross-entropy
on teacher tokens) and KD (the top-$K$ OPD surrogate on both prefix and
suffix), yielding four configurations. \dtdd{} pairs segment-averaged
triggering with BC; the other three change the trigger, suffix loss, or both.
The token-spike controls use a history-normalized z-score trigger.
We additionally evaluate AdaSwitch~\mbox{\citep{adaswitch}} as a separate
baseline, pairing unified KD with its published relative-threshold rule:
takeover occurs when token divergence exceeds three times the mean over the
preceding ten tokens. At step 120, \dtdd{}
reaches $20.83\%$, compared with
$10.23$--$16.79\%$ for the other three and $19.95\%$ for AdaSwitch.
The suffix-loss variants retain the early accuracy gain across weights and
target types (Table~\ref{tab:suffix-objective}). Offline SFT followed by OPD
provides a complementary warm-up baseline. Appendix~\ref{app:base-controls}
reports the individual seeds, trigger details, and controller costs.
Among the online variants in Table~\ref{tab:main-base-controls}, \dtdd{}
achieves the highest accuracy at all four checkpoints. At update 20,
AdaSwitch leads ($16.35\%$ versus $12.65\%$); the two-seed \dtdd{} mean is
higher from update 40 onward (Appendix~\ref{app:base-controls}).

\begin{table*}[t]
  \centering
  \begin{minipage}[t]{0.55\textwidth}
    \caption{Trigger and suffix-loss controls ablations on DAPO-Math-17K.}
    \label{tab:main-base-controls}
  \end{minipage}\hfill
  \begin{minipage}[t]{0.38\textwidth}
    \caption{Teacher-suffix target and weight ablations.}
    \label{tab:suffix-objective}
  \end{minipage}
  \par
  \begin{minipage}[t]{0.55\textwidth}
    \vspace{0pt}
    \centering
    \scriptsize
    \setlength{\tabcolsep}{2.4pt}
\newcommand{\seedstat}[2]{\ensuremath{#1_{\pm #2}}}
\newcommand{\controlrow}{\rule[-2.5pt]{0pt}{10pt}}
\begin{tabularx}{\linewidth}{@{}ll*{4}{>{\raggedleft\arraybackslash}X}r@{}}
\toprule
Method & Suffix loss & \multicolumn{4}{c}{Optimizer step} & Tok. \\
\cmidrule(l){3-6}
 & & 60 & 80 & 100 & 120 & (M) \\
\midrule
\controlrow OPD & -- & 12.14 & 14.00 & 13.15 & 14.54 & 0 \\
\rowcolor{dtddrow}
\controlrow \dtdd{} & BC & 20.32 & 21.47 & 21.08 & 20.83 & 0.52 \\
\controlrow Token spike & KD & 12.47 & 12.14 & 12.81 & 13.51 & 0.40 \\
\controlrow Token spike & BC & 13.19 & 13.07 & 13.26 & 16.79 & 0.43 \\
\controlrow Segment & KD & 5.78 & 7.80 & 8.04 & 10.23 & 0.67 \\
\controlrow AdaSwitch & KD & 18.16 & 19.57 & 18.68 & 19.95 & 0.86 \\
\midrule
\controlrow SFT + OPD & -- & 20.63 & 22.81 & 22.45 & 22.03 & 0$^{\ddagger}$ \\
\bottomrule
\end{tabularx}

    \par\vspace{3pt}
    \raggedright
    Accuracy is average@16 (\%). OPD and \dtdd{} report two-seed means;
    other methods use one run. Triggers: segment-averaged (\dtdd{}/Segment),
    z-score (Token spike), or AdaSwitch's relative-threshold rule.
    BC: behavior cloning on teacher tokens.
    KD: the top-$K$ OPD surrogate on both prefix and suffix.
    Tok.: teacher-written tokens per update,
    in millions. $^{\ddagger}$SFT uses an additional 95M teacher-written
    tokens offline.
  \end{minipage}\hfill
  \begin{minipage}[t]{0.38\textwidth}
    \vspace{0pt}
    \centering
    \scriptsize
    \setlength{\tabcolsep}{2.4pt}
\newcommand{\suffixrow}{\rule[-2.5pt]{0pt}{10pt}}
\begin{tabularx}{\linewidth}{@{}lc*{4}{>{\raggedleft\arraybackslash}X}@{}}
\toprule
Suffix loss & $\lambda_{\mathrm{BC}}$ & \multicolumn{4}{c}{Optimizer step} \\
\cmidrule(l){3-6}
 & & 60 & 80 & 100 & 120 \\
\midrule
\rowcolor{dtddrow}
\suffixrow BC & 1 & 20.16 & 21.34 & 21.25 & 20.76 \\
\addlinespace[4pt]
\multicolumn{6}{@{}l}{\suffixrow\textit{Loss-weight ablation}} \\
\suffixrow BC & 3 & 21.15 & 21.48 & 19.91 & 19.10 \\
\suffixrow BC & 0.1 & 20.94 & 20.31 & 22.40 & 22.29 \\
\addlinespace[4pt]
\multicolumn{6}{@{}l}{\suffixrow\textit{Target ablation}} \\
\suffixrow Soft CE & 1 & 21.27 & 22.01 & 19.95 & 20.59 \\
\bottomrule
\end{tabularx}

    \par\vspace{3pt}
    \raggedright
    Accuracy is average@16 (\%), one run per variant.
    BC: behavior cloning with one-hot cross-entropy on teacher tokens.
    Soft CE: cross-entropy with soft teacher targets.
    $\lambda_{\mathrm{BC}}$ weights the suffix loss. The highlighted
    default uses \dtdd{} seed 1; Table~\ref{tab:main-base-controls}
    reports the two-seed mean.
  \end{minipage}
  \vspace{-5mm}
\end{table*}

\subsection{Q1: Does \dtdd{} Stabilize Long-Horizon OPD?}
\label{sec:stability}

Long generation budgets expose a late OPD instability that \dtdd{} suppresses.
We hold the student, teacher, data, sampling, and optimizer fixed while varying
the maximum rollout length over 7,168, 15,360, and 18,432 tokens. We quantify
late-training instability by the maximum gradient norm from update 51 onward,
after the short-lived optimization transient at the start of training.

The 7K control remains stable under both methods, with peak gradient norms of
$0.54$ for OPD and $0.96$ for \dtdd{}. At 15K, OPD enters a sharp late event:
its gradient norm reaches $12.02$ and its truncated-rollout ratio reaches
$52\%$. \dtdd{} limits the same quantities to $1.40$ and $32\%$, reducing the
gradient peak by $8.6\times$. The 18K run reproduces the pattern, with gradient
peaks of $12.81$ for OPD and $1.84$ for \dtdd{} ($7.0\times$). Held-out
accuracy remains within one point across all three budgets. Thus, \dtdd{}
removes the long-budget optimization failure while preserving the performance
level reached by OPD. Appendix~\ref{app:stability} provides the complete
entropy, truncation, overlap, reward, and accuracy trajectories.

Clipping OPD's token-level advantages to $[-1,1]$ also suppresses the late
spikes at both long budgets, with held-out accuracy within 1.1 points of
the unclipped reproduction (Appendix~\ref{app:stability-clipping}).
\begin{figure*}[!h]
  \centering
    \includegraphics[width=\textwidth]{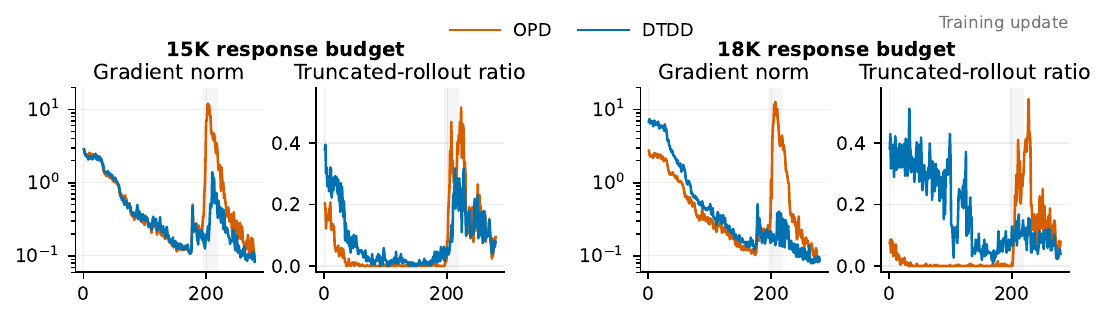}

  \caption{\dtdd{} suppresses late OPD spikes on DAPO-Math-17K at 15K/18K response budgets.}
  \label{fig:stability}
  \vspace{-6mm}
\end{figure*}

\subsection{Q2: Can Takeover Warm Up a Base Student Online?}
\label{sec:efficiency}

\begin{figure}[t]
  \centering
  \includegraphics[width=0.90\columnwidth]{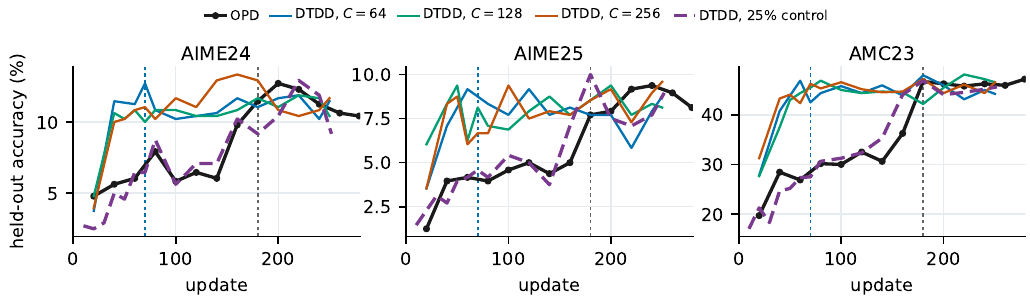}
  \caption{Held-out accuracy for Qwen3-1.7B-Base $\leftarrow$
  Qwen3-4B-Base-GRPO on DAPO-Math-17K. The length-gated control starts near
  $25\%$ takeover and adapts thereafter.}
  \label{fig:efficiency}
\end{figure}

\dtdd{} can warm up a base student online, without first preparing an SFT
dataset and warm-starting the student on it. Figure~\ref{fig:efficiency}
compares a Qwen3-1.7B-Base student with a Qwen3-4B-Base-GRPO teacher on
DAPO-Math-17K. High-dose \dtdd{} reaches the $21.3\%$ held-out reference level
near update 70, whereas OPD reaches it near update 180. The trajectories later
converge near $22\%$ average accuracy. The time to match OPD's update-240
accuracy varies across window sizes (Appendix Table~\ref{tab:app-crossing}).
Appendix~\ref{app:base-results}
reports the complete dose, window-size, and cost analyses.

We measure intervention frequency by the \emph{takeover rate}: the fraction of
sampled rollouts in which divergence triggers a teacher takeover.
\emph{Rewind} replaces the divergent suffix; \emph{restart} replaces the
entire rollout. Fixed thresholds allow takeover rates to vary with divergence,
whereas target-rate controllers adjust $\tau$ to maintain a prescribed rate.

Figure~\ref{fig:efficiency} compares OPD with four rewind configurations.
Three form a window-size ablation: $C\in\{64,128,256\}$, with fixed thresholds
$\tau\in\{0.475,0.362,0.296\}$ calibrated to match an initial takeover rate of
approximately $90\%$. The fourth fixes $\tau=0.2$ at $C=256$ and makes the
third segment ($j_{\min}=3$) the first eligible for takeover. This control
starts near $25\%$ takeover, providing limited teacher supervision early in
training. As model rollouts become longer, more of them satisfy the trigger;
the resulting increase in takeover rate supplies additional teacher
supervision.

At update 120, the three high-initial-takeover arms reach $21.32\%$ ($C=64$),
$20.90\%$ ($C=128$), and $21.23\%$ ($C=256$), whereas the fixed-$\tau=0.2$
control remains close to OPD ($14.81\%$ versus $14.65\%$). A separate
rate-controlled restart setting targeting $45\%$ takeover reaches $20.49\%$.
By update 240, all four rewind configurations reach $21.04$--$22.40\%$,
comparable to OPD at $22.29\%$. Thus, for a base student without prior SFT
warm-up, sufficient coherent teacher continuations at a high initial dose
accelerate online warm-up; segment-level divergence routes them online and lets
fixed-threshold intervention recede with alignment.

\begin{figure}[t]
  \centering  \includegraphics[width=0.90\columnwidth]{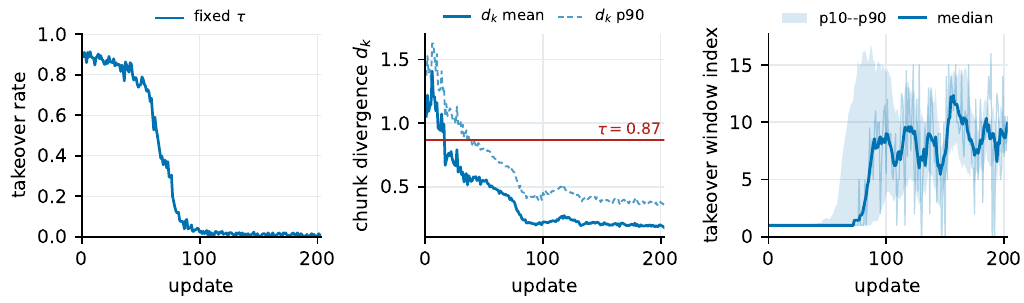}
  \caption{Fixed-threshold dynamics ($\tau=0.87$) for Qwen3-1.7B-Base
  $\leftarrow$ Qwen3-4B on OpenThoughts3.}
  \label{fig:controller}
\end{figure}

The three high-coverage DAPO arms---excluding the low-initial-takeover
$\tau=0.2$ control---show decreasing segment divergence and teacher-written
token share, together with later takeover positions (Appendix
Figure~\ref{fig:app-dapo-dynamics}). Figure~\ref{fig:controller} illustrates
these dynamics on OpenThoughts3.

With $\tau=0.87$, the OpenThoughts3 takeover rate falls from $88.3\%$ to below
$5\%$ by update 87 and to zero by update 150, while the median trigger moves
from window 1 to 10.5. Among rollouts that still trigger, the later position
retains a longer student prefix before teacher continuation. At matched update
200, the three OpenThoughts3 controllers reach $14.36$--$15.75\%$ accuracy,
compared with $12.85\%$ for OPD
(Appendix Table~\ref{tab:app-matched-endpoint}).

On DAPO-Math-17K, \dtdd{} reaches $21.34\%$ held-out accuracy in 22.0 hours;
offline SFT followed by OPD reaches $21.55\%$ in approximately 42.8 serial
hours, including teacher-data generation and SFT. \dtdd{} uses 43M versus 95M
teacher-written tokens, allocated online within one training stage.
Appendix~\ref{app:base-cost} reports the checkpoint-level timing and hardware
details.

\subsection{Q3: Does Divergence-Guided Routing Transfer to Agentic Tasks?}
\label{sec:agentic-results}

In multi-turn tasks, \dtdd{} triggers teacher recovery at action boundaries,
where each action contains a reasoning span and tool call. Takeover is used
only during training; evaluation uses the student alone.

On ALFWorld, we pair a LoRA-adapted Qwen2.5-3B-Instruct student with the
ALFWorld-specific GiGPO-7B teacher and set $\tau=0.7$. As in the reasoning
experiments, training begins with high teacher coverage ($90.2\%$ takeover).
As the student improves over training, takeover falls to near zero, automatically
reducing teacher supervision. Near update 180, a transient rise in discrepancy
reactivates takeover despite its previously near-zero rate (Appendix
Figure~\ref{fig:agentic-dynamics}). \dtdd{} therefore also acts as an online
monitor and recovery mechanism: it remains mostly inactive after alignment but
resumes corrective teacher supervision when instability emerges. With
checkpoints evaluated every 20 updates, \dtdd{} leads OPD at all ten matched
checkpoints after update 200 in this run. Best-observed average success
increases from $76.49\%$ to $81.32\%$, with the largest gain on the unseen
split (Table~\ref{tab:main-results}). Appendix~
\ref{app:alfworld-intervention} reports the full trajectory and adaptation
controls, including the matched update-400 comparison.

The agentic code experiment examines intervention dose for an SFT-initialized
policy, pairing Qwen3-1.7B-SFT with DemyAgent-GRPO-4B on the code-only subset
of Open-AgentRL-30K \citep{demyagent}. Our main setting uses $\tau=0.50$,
yielding a sparse $2.8\%$ teacher-token dose. At their best-observed
checkpoints, sparse takeover improves all four evaluation splits and raises
aggregate accuracy from $36.49\%$ to $40.59\%$ (Table~\ref{tab:main-results}).
This result complements the base-student study: high coverage supplies a
missing reasoning pattern, whereas sparse correction preserves the on-policy
interaction of an already viable agent. Appendix~\ref{app:agentic-code-dose}
reports matched endpoints and the dose ablation.

\begin{table*}[t]
  \centering
  \caption{Best-observed results (\%; selection details in
  Appendix~\ref{app:evaluation-reporting}). Subscripts show percentage-point
  changes from OPD; bold marks the best trained method.}
  \label{tab:main-results}
  \scriptsize
  \setlength{\tabcolsep}{4pt}
\begin{tabular}{@{}p{0.28\textwidth}rrrrr@{}}
\toprule
\rowcolor{settingrow}
\multicolumn{6}{l}{\textbf{Math reasoning:} student Qwen3-1.7B-Base; teacher Qwen3-4B (non-thinking); data OpenThoughts3} \\
Method & AIME24 & AIME25 & AMC23 & & Average $\uparrow$ \\
\cmidrule(lr){1-1}\cmidrule(lr){2-4}\cmidrule(lr){6-6}
\rowcolor{teacherrow}
Teacher & \emph{21.04} & \emph{20.62} & \emph{67.66} & & \emph{36.44} \\
OPD & 5.62 & 5.42 & 27.50 & & 12.85 \\
\rowcolor{dtddrow}
\dtdd{} fixed $\tau=0.87$ & 6.88\posdiff{1.26} & \best{7.08}\posdiff{1.66} & 31.41\posdiff{3.91} & & 15.12\posdiff{2.27} \\
\rowcolor{dtddrow}
\dtdd{} rewind, target $90\%\!\rightarrow\!5\%$ & 7.29\posdiff{1.67} & 4.79\negdiff{0.63} & 32.19\posdiff{4.69} & & 14.76\posdiff{1.91} \\
\rowcolor{dtddrow}
\dtdd{} restart, target $90\%\!\rightarrow\!5\%$ & \best{8.75}\posdiff{3.13} & 5.21\negdiff{0.21} & \best{33.28}\posdiff{5.78} & & \best{15.75}\posdiff{2.90} \\
\addlinespace

\rowcolor{settingrow}
\multicolumn{6}{l}{\textbf{Agentic interaction:} student Qwen2.5-3B-Instruct (LoRA); teacher GiGPO-7B; data ALFWorld} \\
Method & Seen & Unseen & & & Average $\uparrow$ \\
\cmidrule(lr){1-1}\cmidrule(lr){2-3}\cmidrule(lr){6-6}
\rowcolor{teacherrow}
Teacher & \emph{85.71} & \emph{88.06} & & & \emph{86.89} \\
OPD & 83.57 & 69.40 & & & 76.49 \\
\rowcolor{dtddrow}
\dtdd{} & \best{84.29}\posdiff{0.72} & \best{78.36}\posdiff{8.96} & & & \best{81.32}\posdiff{4.83} \\
\addlinespace

\rowcolor{settingrow}
\multicolumn{6}{l}{\textbf{Agentic code:} student Qwen3-1.7B-SFT; teacher DemyAgent-GRPO-4B; data Open-AgentRL-30K (code only)} \\
Method & HO-200 full & HO-200 sub & LCB v1 & LCB v6 & Average $\uparrow$ \\
\cmidrule(lr){1-1}\cmidrule(lr){2-5}\cmidrule(lr){6-6}
\rowcolor{teacherrow}
Teacher & \emph{62.12} & \emph{64.19} & \emph{68.78} & \emph{47.43} & \emph{60.63} \\
OPD & 31.00 & 30.56 & 50.91 & 33.48 & 36.49 \\
\rowcolor{dtddrow}
\dtdd{} & \best{35.81}\posdiff{4.81} & \best{37.62}\posdiff{7.06} & \best{53.65}\posdiff{2.74} & \best{35.27}\posdiff{1.79} & \best{40.59}\posdiff{4.10} \\
\bottomrule
\end{tabular}

\end{table*}

\subsection{Q4: What Determines Whether Teacher Takeover Helps?}
\label{sec:boundaries}

\begin{figure*}[t]
  \centering
  \includegraphics[width=0.8\textwidth]{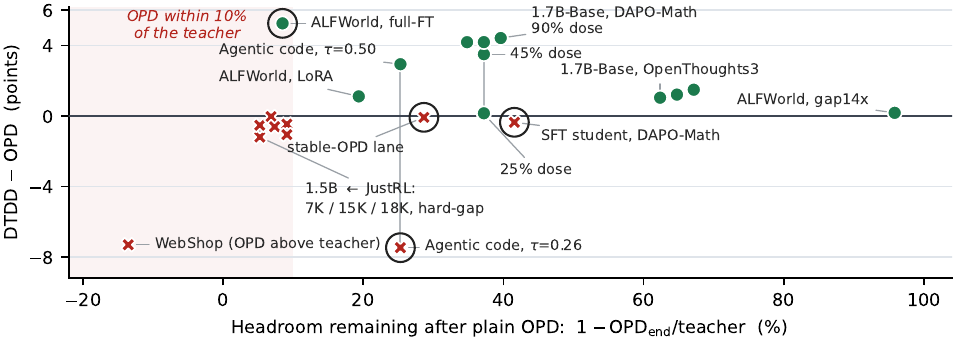}
  \caption{\dtdd{} minus OPD, averaged over all matched checkpoints, against
  the headroom plain OPD leaves below its teacher. Points left of zero are
  settings where OPD ends above the teacher's own score. Ringed markers identify
  arms analyzed by the validity, learnability, and dose controls in
  Appendix~\ref{app:boundary-results}.}
  \label{fig:headroom}
\end{figure*}

Headroom is an empirical diagnostic of the observed regimes. We define it as
$h=1-\mathrm{OPD}_{\mathrm{end}}/\mathrm{teacher}$, with both scores measured
on the same evaluation harness. Across 23 arms from 13 settings, 19 follow a
common pattern: gains concentrate in regimes where OPD retains more headroom
below the teacher (Figure~\ref{fig:headroom}).

Three controls refine this diagnostic. In the rebuilt WebShop environment, the
task-specific teacher falls below OPD, so takeover routes training toward an
invalid supervision source. Conversely, the 0.5B ALFWorld student retains
substantial teacher headroom but does not reliably learn the recovery, exposing
a student-learnability constraint. Finally, the agentic-code dose ablation
shows that excessive intervention can replace too much on-policy interaction.
Appendix~\ref{app:boundary-results} reports the complete settings,
per-arm results, and TCOD comparison.

\section{Conclusion and Limitations}
\label{sec:conclusion}

\dtdd{} uses segment-level student--teacher divergence to route supervision
between standard OPD and teacher-generated recovery. Across paired experiments,
sparse takeover suppresses
long-budget optimization extremes, high initial teacher coverage accelerates
learning from a base initialization, and action-level takeover improves
ALFWorld and agentic code. Controlled comparisons favor segment-level
triggering with behavior cloning. Relative to offline SFT followed by OPD,
this single-stage approach reaches comparable accuracy while cutting
measured serial time by a factor of $1.9$ and using $55\%$ fewer teacher-written
tokens. DTDD uses manually specified thresholds or target takeover rates.
Its divergence trigger does not assess whether teacher recovery is useful for
the student. Automatic dose calibration and recovery-quality assessment are
directions for future work.

\newpage

\bibliography{iclr2027_conference}

\clearpage
\appendix
\suppressfloats[t]

\section*{Appendix}
\begingroup
\small
\setlength{\tabcolsep}{3pt}
\renewcommand{\arraystretch}{1.08}
\noindent\begin{tabular}{@{}p{0.84\columnwidth}r@{}}
\hyperref[app:opd-limitations]{\textbf{\ref*{app:opd-limitations}}\quad Measured Limitations of Vanilla OPD}
  & \pageref*{app:opd-limitations} \\
\hyperref[app:experimental-details]{\textbf{\ref*{app:experimental-details}}\quad Experimental Details and Reporting Protocols}
  & \pageref*{app:experimental-details} \\
\hspace{1em}\hyperref[app:reasoning-protocol]{\ref*{app:reasoning-protocol}\quad Reasoning Training Protocol}
  & \pageref*{app:reasoning-protocol} \\
\hspace{1em}\hyperref[app:interactive-protocol]{\ref*{app:interactive-protocol}\quad Interactive Training Protocol}
  & \pageref*{app:interactive-protocol} \\
\hspace{1em}\hyperref[app:evaluation-reporting]{\ref*{app:evaluation-reporting}\quad Evaluation and Checkpoint Selection}
  & \pageref*{app:evaluation-reporting} \\
\hspace{1em}\hyperref[app:matched-comparisons]{\ref*{app:matched-comparisons}\quad Matched-Checkpoint Comparisons}
  & \pageref*{app:matched-comparisons} \\
\hyperref[app:stability]{\textbf{\ref*{app:stability}}\quad Additional Long-Horizon Stability Results}
  & \pageref*{app:stability} \\
\hspace{1em}\hyperref[app:stability-dynamics]{\ref*{app:stability-dynamics}\quad Training Dynamics across Response Budgets}
  & \pageref*{app:stability-dynamics} \\
\hspace{1em}\hyperref[app:stability-clipping]{\ref*{app:stability-clipping}\quad Signal-Clipping Control}
  & \pageref*{app:stability-clipping} \\
\hspace{1em}\hyperref[app:stability-controller]{\ref*{app:stability-controller}\quad Intervention Dynamics and Exploration-Floor Ablation}
  & \pageref*{app:stability-controller} \\
\hyperref[app:base-results]{\textbf{\ref*{app:base-results}}\quad Additional Base-Initialization Results}
  & \pageref*{app:base-results} \\
\hspace{1em}\hyperref[app:base-dose-response]{\ref*{app:base-dose-response}\quad Learning Curves and Matched-Checkpoint Results}
  & \pageref*{app:base-dose-response} \\
\hspace{1em}\hyperref[app:base-controls]{\ref*{app:base-controls}\quad Controller Dynamics and Design Ablations}
  & \pageref*{app:base-controls} \\
\hspace{1em}\hyperref[app:base-cost]{\ref*{app:base-cost}\quad Offline SFT Baseline and Computational Cost}
  & \pageref*{app:base-cost} \\
\hyperref[app:agentic-results]{\textbf{\ref*{app:agentic-results}}\quad Additional Agentic Results}
  & \pageref*{app:agentic-results} \\
\hspace{1em}\hyperref[app:alfworld-intervention]{\ref*{app:alfworld-intervention}\quad ALFWorld Action-Level Intervention}
  & \pageref*{app:alfworld-intervention} \\
\hspace{1em}\hyperref[app:agentic-code-dose]{\ref*{app:agentic-code-dose}\quad Agentic Code Dose}
  & \pageref*{app:agentic-code-dose} \\
\hyperref[app:boundary-results]{\textbf{\ref*{app:boundary-results}}\quad Headroom Analysis and Boundary Cases}
  & \pageref*{app:boundary-results} \\
\hspace{1em}\hyperref[app:headroom-inventory]{\ref*{app:headroom-inventory}\quad Headroom Inventory and Reporting Details}
  & \pageref*{app:headroom-inventory} \\
\hspace{1em}\hyperref[app:headroom-boundaries]{\ref*{app:headroom-boundaries}\quad Boundary Cases}
  & \pageref*{app:headroom-boundaries} \\
\hspace{1em}\hyperref[app:tcod-reanalysis]{\ref*{app:tcod-reanalysis}\quad Reanalysis of TCOD Results}
  & \pageref*{app:tcod-reanalysis} \\
\end{tabular}
\endgroup

\clearpage
\suppressfloats[t]
\section{Measured Limitations of Vanilla OPD}
\label{app:opd-limitations}

\begin{figure*}[!htbp]
  \centering
  \includegraphics[width=\textwidth]{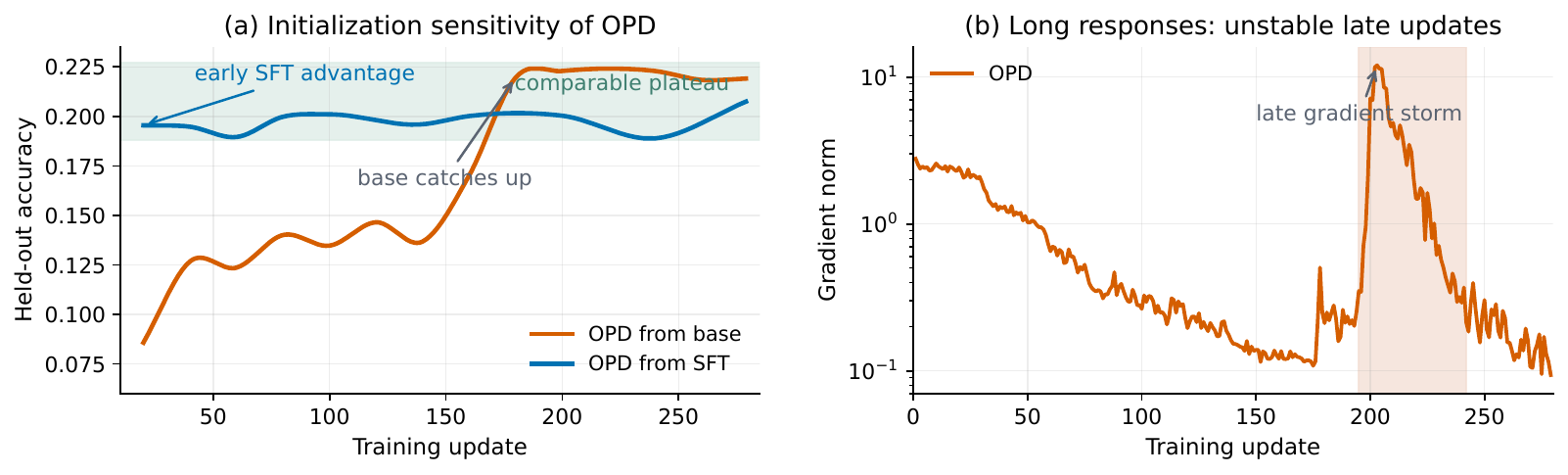}
  \caption{Two representative limitations of vanilla OPD. \emph{Left:}
  Qwen3-1.7B-Base and Qwen3-1.7B-SFT are distilled from the same
  Qwen3-4B-Base-GRPO teacher on DAPO-Math-17K with a 7,168-token generation
  budget. The teacher, data, optimizer, and generation budget are held fixed;
  the SFT initialization has a large early advantage, while the base
  initialization eventually reaches a
  comparable plateau. Held-out accuracy is the mean avg@16 score on AIME 2024,
  AIME 2025, and AMC 2023. \emph{Right:}
  DeepSeek-R1-Distill-Qwen-1.5B is distilled from JustRL-DeepSeek-1.5B on
  DAPO-Math-17K with a 15,360-token generation budget. OPD develops a large late
  gradient spike despite otherwise decreasing gradient norms.}
  \label{fig:opd-limitations}
  \vspace{-3mm}
\end{figure*}

Figure~\ref{fig:opd-limitations} supplements the schematic failures in
Figure~\ref{fig:opd-failure-cases}(A--B) with measured OPD learning curves:
a controlled comparison of base and SFT initializations, and the gradient
trajectory under a long response budget.

\section{Experimental Details and Reporting Protocols}
\label{app:experimental-details}

This appendix expands Section~\ref{sec:setup} with task-specific generation
budgets (Table~\ref{tab:setup}), optimization and controller settings,
interactive training objectives, and evaluation protocols.

\begin{table*}[!htbp]
  \centering
  \caption{Experimental settings. The budget specifies the maximum generated
  tokens for reasoning tasks and the maximum number of environment or
  tool turns for interactive tasks.}
  \label{tab:setup}
  \footnotesize
  \setlength{\tabcolsep}{3pt}
\begin{tabular}{@{}p{0.115\textwidth}p{0.18\textwidth}p{0.18\textwidth}p{0.165\textwidth}p{0.14\textwidth}p{0.13\textwidth}@{}}
\toprule
Setting & Student & Teacher & Training data & Budget & Evaluation \\
\midrule
Rollout length
  & R1-Distill-Qwen-1.5B & JustRL-DeepSeek-1.5B
  & DAPO-Math-17K & 7,168 / 15,360 / 18,432 tokens
  & AIME24/25, AMC23 \\
Base student
  & Qwen3-1.7B-Base & Qwen3-4B-Base-GRPO
  & DAPO-Math-17K & 7,168 tokens
  & AIME24/25, AMC23 \\
Base student
  & Qwen3-1.7B-Base & Qwen3-4B (non-thinking)
  & OpenThoughts3 & 7,168 tokens
  & AIME24/25, AMC23 \\
ALFWorld
  & Qwen2.5-3B-Instruct & GiGPO-7B
  & ALFWorld train & 30 turns; 512 tokens/action
  & Seen / unseen success \\
WebShop
  & Qwen2.5-3B-Instruct & GiGPO-7B (WebShop)
  & Rebuilt 120K catalog & 15 turns; 512 tokens/action
  & Success \\
Agentic code
  & Qwen3-1.7B-SFT & DemyAgent-GRPO-4B
  & Open-AgentRL-30K (code only) & 8 turns / 12,288 tokens
  & LCB v1/v6, HO-200 full/sub \\
\bottomrule
\end{tabular}

\end{table*}

\subsection{Reasoning Training Protocol}
\label{app:reasoning-protocol}

The reasoning comparisons in Sections~\ref{sec:stability}
and~\ref{sec:efficiency} use the shared training protocol below;
Table~\ref{tab:app-reasoning-config} specifies the controller for each arm. Each
update samples 64 prompts with four rollouts per prompt. Student and
teacher-logit temperatures are both 1.0, and each rollout batch receives one
policy update. Optimization uses AdamW with
$(\beta_1,\beta_2)=(0.9,0.999)$, $\epsilon=10^{-8}$, weight decay $0.01$,
and global gradient clipping at $1.0$. The learning rate is held constant at
$10^{-6}$ without warm-up. Distillation uses the student top-$K$ set
with $K=16$, student-probability weighting, and token-mean aggregation. The
teacher is frozen; no sequence-level outcome reward enters the objective, and
teacher-written tokens use $\lambda_{\mathrm{BC}}=1$ unless varied in the
suffix-loss ablation. The length sweep and length-gated $\tau=0.2$ control set
$j_{\min}=3$; the other base-initialization configurations use $j_{\min}=1$. The
256-token continuation-budget check does not filter any reported trigger.

Both terms in Equation~\ref{eq:dtdd-objective} share the same optimization
path. The prefix uses the clipped policy-gradient form of
$\widehat\ell_t^{\mathrm{OPD}}$; the suffix uses a one-hot target, which gives
the standard behavior-cloning gradient when clipping is inactive.

Reasoning takeover is retrospective. The student first generates a full
rollout, after which \dtdd{} computes segment discrepancies and requests a
teacher continuation for each triggered example. The update retains the
student prefix and excludes the replaced student suffix. This pipeline changes
the training sequence without reducing student rollout compute, so
Section~\ref{sec:efficiency} measures optimization-step efficiency.

\subsubsection{Controller and Recovery Details}
\label{app:controller-details}

For rollout-level rewind, we suppress triggers too close to the maximum
generation budget to leave at least $B_{\min}$ tokens for teacher continuation.
We instantiate $\tau_s$ with either a fixed or rate-controlled threshold. The
fixed controller holds $\tau_s=\tau$, allowing takeover to recede as
student--teacher discrepancy falls. The rate-controlled variant sets $\tau_s$
to the empirical $(1-r_s)$-quantile of per-rollout maximum eligible segment
discrepancies across the batch, where the target takeover rate $r_s$ may be fixed or
scheduled. Both variants use divergence to select which rollouts trigger
and the first crossing to determine the intervention boundary. In the length
sweep, an exploration branch triggers a fraction $\epsilon=0.05$ of rollouts at a
uniformly sampled candidate boundary as an exploration floor. Once segment
discrepancy contracts below $\tau$, this floor rather than the divergence
signal accounts for most interventions; Section~\ref{app:stability} reports
where each length-sweep arm sits relative to the floor. Unless otherwise
specified, \dtdd{} uses rewind; we consider full-rollout restart ($c=0$) only
as an ablation.

\begin{table*}[t]
  \centering
  \caption{Reasoning configurations. Target rate is a rollout-level control
  parameter; the realized rate and teacher-token dose are measured from the
  generated batches. A dash denotes vanilla OPD.}
  \label{tab:app-reasoning-config}
  \footnotesize
  \setlength{\tabcolsep}{3.5pt}
\begin{tabular}{@{}p{0.165\textwidth}p{0.13\textwidth}p{0.12\textwidth}p{0.12\textwidth}p{0.17\textwidth}p{0.195\textwidth}@{}}
\toprule
Study & Budget & Segment & Mode & Controller & Initialization / data \\
\midrule
Length sweep
  & 7,168 / 15,360 / 18,432
  & 256 tokens
  & rewind
  & fixed $\tau=0.7$, $\epsilon=0.05$
  & DeepSeek-R1-Distill-Qwen-1.5B; DAPO-Math-17K \\
DAPO base, low initial coverage
  & 7,168
  & 256 tokens
  & rewind
  & fixed $\tau=0.2$; initial takeover $\approx25\%$
  & Qwen3-1.7B-Base; DAPO-Math-17K \\
DAPO base, medium coverage
  & 7,168
  & 256 tokens
  & restart
  & target rate $45\%$
  & Qwen3-1.7B-Base; DAPO-Math-17K \\
DAPO base, high initial coverage
  & 7,168
  & 64 / 128 / 256 tokens
  & rewind
  & fixed $\tau=.475/.362/.296$; initial takeover $\approx90\%$
  & Qwen3-1.7B-Base; DAPO-Math-17K \\
OpenThoughts3 base
  & 7,168
  & 256 tokens
  & rewind / restart
  & target rate $90\%\!\rightarrow\!5\%$, or fixed $\tau=0.87$
  & Qwen3-1.7B-Base; OpenThoughts3 \\
\bottomrule
\end{tabular}

\end{table*}

\subsection{Interactive Training Protocol}
\label{app:interactive-protocol}

This section specifies the normalized action-token objectives and optimization
settings for the interactive realization in Section~\ref{sec:trigger}. Let
$z_j\in\{S,T\}$ denote whether the student or teacher writes action $j$, let
$W_j$ contain its generated-token positions, and let $N_S$ and $N_T$ count
student- and teacher-written tokens. Let $a_{j,t}$ be token $t$ of the action
executed at turn $j$, let $h_{j,t}$ be the interaction history and action prefix
before that token, and let $\pi_{\mathrm{old}}$ denote the student that
collected the interaction. The sampled-token OPD signal is
\begin{equation}
  A^{\mathrm{act}}_{j,t}
  = \log \pi_T(a_{j,t}\mid h_{j,t})
    - \log \pi_{\mathrm{old}}(a_{j,t}\mid h_{j,t}).
  \label{eq:agentic-advantage}
\end{equation}
For student-written actions, define the unclipped sampled-token OPD term
$\widehat\ell^{\mathrm{OPD}}_{j,t}
=-\operatorname{sg}[A^{\mathrm{act}}_{j,t}]
\log\pi_\theta(a_{j,t}\mid h_{j,t})$.
The implementations differ in the recovery horizon and the loss applied to
teacher-written actions, as detailed below.

All interactive runs use AdamW with
$(\beta_1,\beta_2)=(0.9,0.999)$, $\epsilon=10^{-8}$, weight decay $0.01$,
a constant learning-rate schedule without warm-up, and global gradient
clipping at $1.0$.

\subsubsection{ALFWorld and WebShop}
\label{app:environment-protocol}

In ALFWorld and WebShop, the teacher replaces the triggering action, and the
student resumes at the next turn. Both action types pass through the same
sampled-token multi-turn OPD surrogate. With $N=N_S+N_T$, the objective is
\begin{equation}
  \widehat{\mathcal L}_{\mathrm{env}}(\theta)
  = -\mathbb E\!\left[\frac{1}{N}
      \sum_j\sum_{t\in W_j}
      \operatorname{sg}\!\left[A^{\mathrm{act}}_{j,t}\right]
      \log\pi_\theta(a_{j,t}\mid h_{j,t})\right],
  \label{eq:agentic-env-objective}
\end{equation}
implemented with the clipped surrogate. Student actions are sampled from the
rollout policy; teacher actions are sampled from the teacher, making the latter
term log-ratio-weighted imitation rather than unit-weight cross-entropy.

\paragraph{Teacher-action supervision.}
The ALFWorld and WebShop implementation reuses the framework's existing
sampled-token advantage in Equation~\ref{eq:agentic-advantage}. Its scoring
interface returns log-probabilities for the generated tokens. For a
teacher-written action, we record both the rollout student's and the teacher's
log-probabilities on those same tokens. This shared representation allows
student and teacher actions to pass through the same trainer. Agentic code
uses a separate actor with an explicit behavior-cloning loss
(Equation~\ref{eq:agentic-code-objective}).

On a teacher-written token, a positive $A^{\mathrm{act}}_{j,t}$ encourages the
student to increase its probability; a negative value reverses this direction.
The weight approaches zero as the student and teacher probabilities match.
Thus, the environment objective uses a signed log-probability-gap weight,
whereas behavior cloning assigns each teacher target a fixed positive weight.
Because these tokens are sampled from the teacher, the weighted surrogate is
not an unbiased estimator of either KL direction. We have not compared it
against unit-weight behavior cloning within ALFWorld or WebShop.

ALFWorld pairs a Qwen2.5-3B-Instruct student with GiGPO-7B for episodes of at
most 30 turns and 512 generated tokens per action. A replacement turn includes
both reasoning and action, generated from the same interaction history as the
discarded student turn. We test both LoRA and full-model optimization.
The LoRA setting uses rank
and scale 32 on all linear layers with learning rate $10^{-4}$; full
fine-tuning uses learning rate $10^{-6}$. Both use a training batch size of 32,
training temperature 1.0, and evaluation temperature 0.4. WebShop uses the
same student, LoRA configuration, and WebShop-specific GiGPO-7B teacher on the
rebuilt 120K-product catalog, with at most 15 turns, 512 generated tokens per
action, and $\tau=0.25$.

\subsubsection{Agentic Code}
\label{app:code-protocol}

Our agentic-code implementation builds on the released SOD training stack
\citep{sod}, using the code-only subset of Open-AgentRL-30K and the
execution-based evaluation protocol in Appendix~\ref{app:evaluation-reporting}.
It applies OPD to student-written tokens and behavior cloning to teacher-written
tokens, with role-specific normalization:
\begin{equation}
  \mathcal L_{\mathrm{code}}(\theta)
  = \mathbb E\!\left[
      \frac{1}{N_S}\sum_{j:z_j=S}\sum_{t\in W_j}
        \widehat\ell^{\mathrm{OPD}}_{j,t}(\theta)
      - \frac{\lambda_{\mathrm{BC}}}{N_T}
        \sum_{j:z_j=T}\sum_{t\in W_j}
        \log\pi_\theta(a_{j,t}\mid h_{j,t})
    \right],
  \label{eq:agentic-code-objective}
\end{equation}
where an empty role-specific term is omitted and the OPD term uses its clipped
implementation.

Agentic code pairs Qwen3-1.7B-SFT with DemyAgent-GRPO-4B on the code-only
portion of Open-AgentRL-30K. Each episode allows eight model turns and 12,288
generated tokens. Takeover occurs online at an action boundary, so no future
student tokens are generated after a trigger; the teacher generates all
remaining turns. Fixed thresholds $0.50$ and $0.26$ produce the sparse- and
high-dose conditions. Training uses learning rate $10^{-6}$, a training
batch of 32 prompts with eight rollouts each, a policy minibatch of 16, and
$\lambda_{\mathrm{BC}}=1$. Training and evaluation use temperature 1.0;
evaluation uses top-$p=0.6$ and eight samples per problem.

\subsection{Evaluation and Checkpoint Selection}
\label{app:evaluation-reporting}

The evaluation summary in Section~\ref{sec:setup} is expanded here with
decoding parameters, split sizes (Table~\ref{tab:app-evaluation}), and the
checkpoints selected for Table~\ref{tab:main-results}. Math checkpoints
use a 16,384-token generation budget and 16 samples per problem at temperature
$0.7$ and top-$p=0.95$; accuracy averages over samples and problems. ALFWorld
evaluates the student alone, with no teacher intervention. Execution-harness
choices can substantially affect coding-agent performance \citep{multiharnessrl};
we therefore use the same execution sandbox and decoding parameters for all
code methods.

\begin{table}[t]
  \centering
  \caption{Held-out evaluation protocol.}
  \label{tab:app-evaluation}
  \footnotesize
  \setlength{\tabcolsep}{4pt}
\begin{tabular}{@{}p{0.19\columnwidth}p{0.35\columnwidth}p{0.34\columnwidth}@{}}
\toprule
Domain & Evaluation set & Sampling / metric \\
\midrule
Math & AIME24, AIME25, AMC23
  & 16 samples, $T=0.7$, top-$p=0.95$; average@16 \\
ALFWorld & 140 seen, 134 unseen tasks
  & Student-only environment success \\
Code & LCB v1/v6, HO-200 full/sub
  & 8 samples, $T=1.0$, top-$p=0.6$; execution accuracy \\
\bottomrule
\end{tabular}

\end{table}

Table~\ref{tab:main-results} reports the performance envelope obtained by
selecting each arm's checkpoint by the Average within its block, considering
only checkpoints with results for every benchmark in that block. For
OpenThoughts3, the selected updates are 200 for OPD, 180 for fixed-threshold
\dtdd{}, 180 for scheduled rewind, and 200 for scheduled restart. ALFWorld
selects updates 280 and 140 for OPD and \dtdd{}, respectively; agentic code
selects updates 80 and 60.

Appendix Table~\ref{tab:app-matched-endpoint} evaluates every trained method within a
block at one shared update: 200 for OpenThoughts3, 400 for ALFWorld, and 60 for
agentic code. Stability and boundary analyses use matched checkpoints, and efficiency
analyses use the first evaluated checkpoint that reaches the specified
performance level. Every curve represents one training run; average@$k$ refers
to $k$ evaluation generations. Checkpoints from the same run are correlated
and are not treated as independent training replicates; average@$k$ measures
evaluation-time sampling variation rather than training-seed uncertainty.

Stability statistics use the maximum gradient norm after update 50, applying
one cutoff across generation budgets. Truncation ratio measures the fraction of
rollouts that reach the maximum budget. Takeover rate measures how often an
intervention occurs, while teacher-token dose measures how much of the training
sequence the teacher writes. Takeover-position statistics condition on
triggered examples; each figure specifies whether it reports a mean or median.

\subsection{Matched-Checkpoint Comparisons}
\label{app:matched-comparisons}

Table~\ref{tab:app-matched-endpoint} complements the best-observed results in
Table~\ref{tab:main-results} by evaluating each block at a shared update.
Learning-curve summaries accompany the
long-horizon stability, DAPO base-initialization, and OpenThoughts3 results in
Appendices~\ref{app:stability-dynamics} and~\ref{app:base-dose-response}.

\begin{table*}[t]
  \centering
  \caption{Matched-checkpoint results (\%). All trained methods within each
  block are evaluated at the same update: 200 for math, 400 for ALFWorld, and
  60$^{*}$ for agentic code. Subscripts show the absolute percentage-point change
  from OPD; bold marks the best trained method. For code, Held-out averages the
  two HO-200 splits and LCB averages LiveCodeBench v1 and v6.}
  \label{tab:app-matched-endpoint}
  \scriptsize
  \setlength{\tabcolsep}{4pt}
\begin{tabular}{@{}p{0.28\textwidth}rrrrr@{}}
\toprule
\rowcolor{settingrow}
\multicolumn{6}{l}{\textbf{Math reasoning:} student Qwen3-1.7B-Base; teacher Qwen3-4B (non-thinking); data OpenThoughts3} \\
Method & AIME24 & AIME25 & AMC23 & & Average $\uparrow$ \\
\cmidrule(lr){1-1}\cmidrule(lr){2-4}\cmidrule(lr){6-6}
\rowcolor{teacherrow}
Teacher & \emph{21.04} & \emph{20.62} & \emph{67.66} & & \emph{36.44} \\
OPD & 5.62 & \best{5.42} & 27.50 & & 12.85 \\
\rowcolor{dtddrow}
\dtdd{} fixed $\tau=0.87$ & 6.88\posdiff{1.25} & \best{5.42}\posdiff{0.00} & 31.25\posdiff{3.75} & & 14.51\posdiff{1.67} \\
\rowcolor{dtddrow}
\dtdd{} rewind, target $90\%\!\rightarrow\!5\%$ & \best{8.96}\posdiff{3.33} & 4.58\negdiff{0.83} & 29.53\posdiff{2.03} & & 14.36\posdiff{1.51} \\
\rowcolor{dtddrow}
\dtdd{} restart, target $90\%\!\rightarrow\!5\%$ & 8.75\posdiff{3.13} & 5.21\negdiff{0.21} & \best{33.28}\posdiff{5.78} & & \best{15.75}\posdiff{2.90} \\
\addlinespace

\rowcolor{settingrow}
\multicolumn{6}{l}{\textbf{Agentic interaction:} student Qwen2.5-3B-Instruct (LoRA); teacher GiGPO-7B; data ALFWorld} \\
Method & Seen & Unseen & & & Average $\uparrow$ \\
\cmidrule(lr){1-1}\cmidrule(lr){2-3}\cmidrule(lr){6-6}
\rowcolor{teacherrow}
Teacher & \emph{85.71} & \emph{88.06} & & & \emph{86.89} \\
OPD & 70.71 & 69.40 & & & 70.06 \\
\rowcolor{dtddrow}
\dtdd{} & \best{84.29}\posdiff{13.57} & \best{70.90}\posdiff{1.49} & & & \best{77.59}\posdiff{7.53} \\
\addlinespace

\rowcolor{settingrow}
\multicolumn{6}{l}{\textbf{Agentic code:} student Qwen3-1.7B-SFT; teacher DemyAgent-GRPO-4B; data Open-AgentRL-30K (code only)} \\
Method & Held-out & LCB & & & Average $\uparrow$ \\
\cmidrule(lr){1-1}\cmidrule(lr){2-3}\cmidrule(lr){6-6}
\rowcolor{teacherrow}
Teacher & \emph{63.16} & \emph{58.11} & & & \emph{60.63} \\
OPD & 26.03 & 41.65 & & & 33.84 \\
\rowcolor{dtddrow}
\dtdd{} ($2.8\%$ dose) & \best{36.72}\posdiff{10.69} & \best{44.46}\posdiff{2.81} & & & \best{40.59}\posdiff{6.75} \\
\rowcolor{dtddrow}
\dtdd{} ($\sim50\%$ dose) & 27.31\posdiff{1.28} & 37.43\negdiff{4.22} & & & 32.37\negdiff{1.47} \\
\bottomrule
\end{tabular}

  \par\vspace{3pt}
  \parbox{\linewidth}{\scriptsize $^{*}$Agentic code is compared at update 60
  rather than at its final update, 130. By update 50, OPD's aggregate accuracy
  has stopped improving. It stays between 33.2\% and 36.5\% over updates 50--130
  and is 33.84\% at update 60 versus 33.93\% at update 130, so later checkpoints
  add no OPD gain. Update 60 is also the best checkpoint for sparse \dtdd{}.
  After it, sparse \dtdd{} drifts downward, but it still leads OPD by 1.03
  points at update 130 (Table~\ref{tab:app-code-endpoint}).}
\end{table*}

The learning-curve summaries distinguish endpoint performance from gains
earlier in training. Last $\Delta$ is the \dtdd{}--OPD accuracy difference at
the final update evaluated for both methods. Last-five $\Delta$ averages the
paired differences over the final five shared checkpoints; all-shared
$\Delta$ averages them over all $N$ shared evaluation checkpoints. All
differences are measured in percentage points of held-out average@16 accuracy.
The OPD late range is its maximum minus minimum accuracy over the final five
shared checkpoints. It describes variation within one training run, including
any learning trend over that window.

\section{Additional Long-Horizon Stability Results}
\label{app:stability}

\subsection{Training Dynamics across Response Budgets}
\label{app:stability-dynamics}

\begin{figure*}[!htbp]
  \centering
  \includegraphics[width=0.98\textwidth]{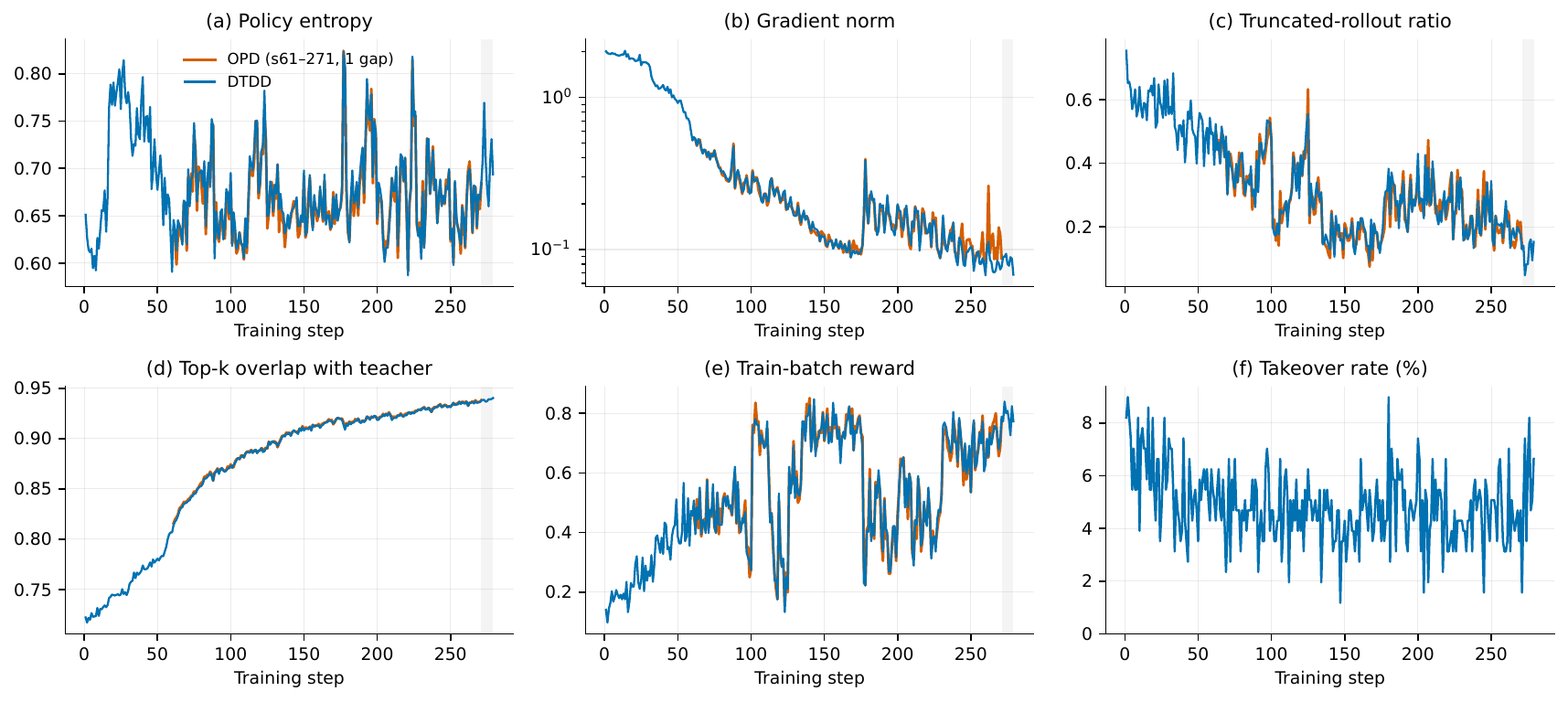}
  \caption{Complete 7K training dynamics for
  DeepSeek-R1-Distill-Qwen-1.5B distilled from JustRL-DeepSeek-1.5B on
  DAPO-Math-17K. This setting serves as the stable control. The OPD telemetry
  begins at update 61; held-out comparisons use checkpoints available for both
  methods.}
  \label{fig:app-storm-7k}
  \vspace{-3mm}
\end{figure*}

\begin{figure*}[t]
  \centering
  \includegraphics[width=0.98\textwidth]{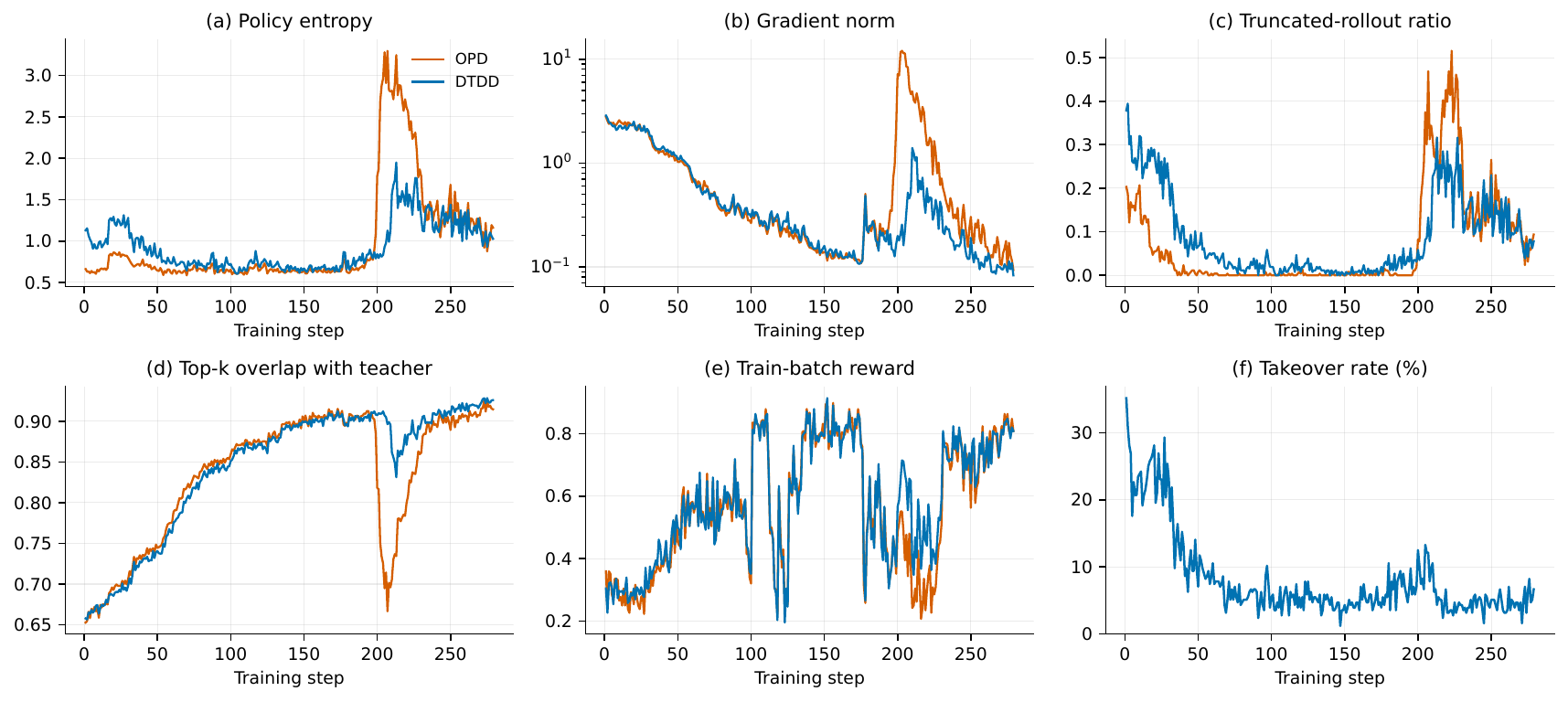}
  \caption{Complete 15K training dynamics. The OPD gradient and entropy spikes
  coincide with increasing truncation and decreasing teacher overlap. \dtdd{}
  limits each component of this event while preserving held-out accuracy.}
  \label{fig:app-storm-15k}
  \vspace{-3mm}
\end{figure*}

\begin{figure*}[t]
  \centering
  \includegraphics[width=0.98\textwidth]{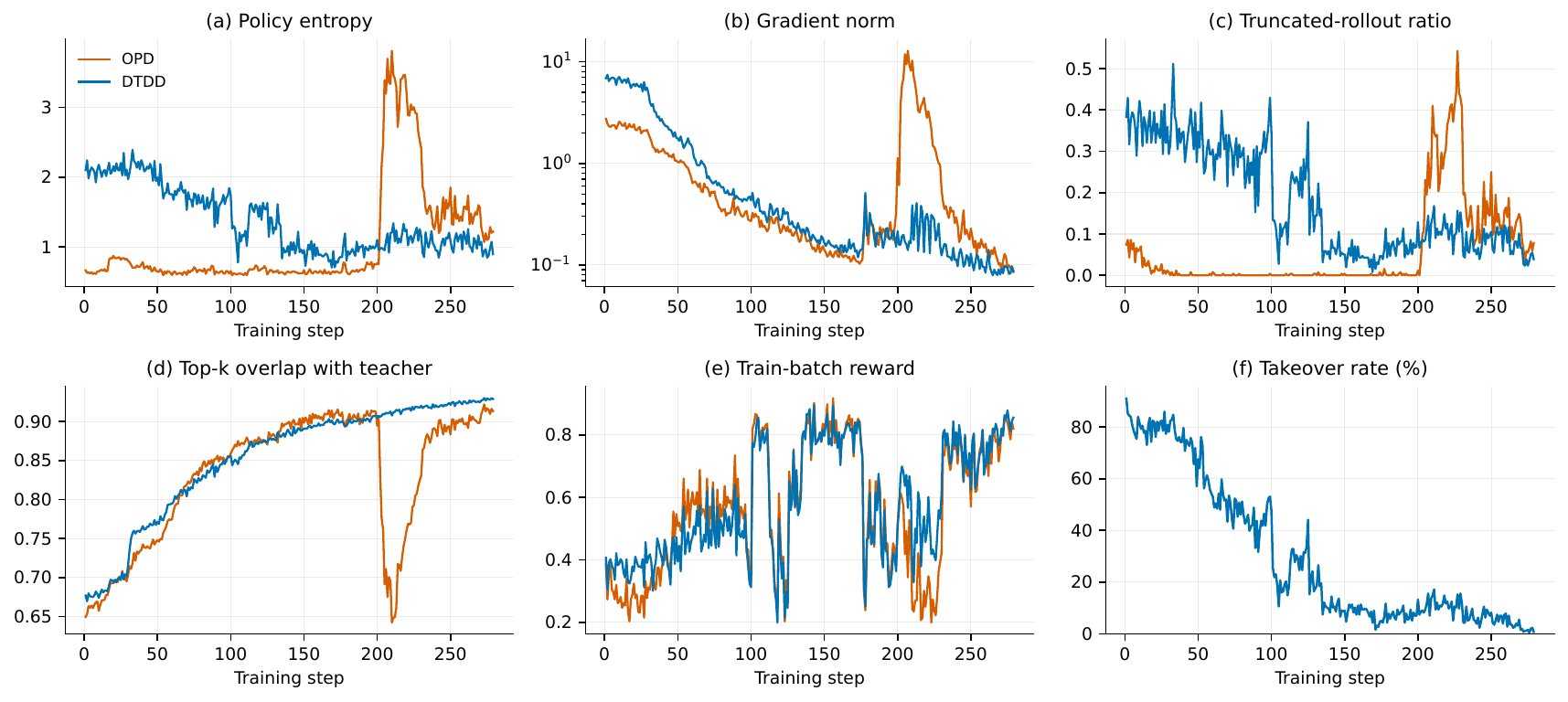}
  \caption{Complete 18K training dynamics. \dtdd{} suppresses the late OPD
  instability in gradient norm, entropy, truncation, and teacher overlap. Its
  intervention rate decreases over training and remains responsive to the
  late discrepancy increase.}
  \label{fig:app-storm-18k}
  \vspace{-3mm}
\end{figure*}

Figures~\ref{fig:app-storm-7k}--\ref{fig:app-storm-18k} extend
Figure~\ref{fig:stability} with a 7K stable control and additional entropy,
teacher-overlap, train-batch reward, and takeover-rate curves. At 7K, OPD
and \dtdd{} follow almost the same entropy, gradient, truncation, teacher-overlap, and
train-batch reward trajectories (Figure~\ref{fig:app-storm-7k}).

At 15K, OPD destabilizes near update 200: gradient norm and entropy rise,
truncation increases, and teacher top-$K$ overlap drops together
(Figure~\ref{fig:app-storm-15k}). These additional metrics show that the
gradient event in Figure~\ref{fig:stability} coincides with changes in
rollout behavior, not just the update magnitude.

The 18K run reproduces the coupled event (Figure~\ref{fig:app-storm-18k}),
with entropy and truncation rising as teacher overlap falls. These runs have
limited accuracy headroom: the JustRL teacher scores $58.45\%$ and OPD reaches
$54.2$--$55.4\%$, leaving $h=5.3$--$7.4\%$, the smallest positive headroom
among the 23 arms in Section~\ref{sec:boundaries}
(Table~\ref{tab:app-headroom-inventory}). In this regime, the benefit of
\dtdd{} is optimization stability rather than higher accuracy.

The matched-curve summaries give the same accuracy picture: the final-five
\dtdd{}--OPD differences are $-0.73$ and $-0.20$ points at 15K and 18K,
respectively. Averaging over all 10 and 14 shared checkpoints through update
279 gives $-0.62$ and $-1.20$ points (definitions in
Appendix~\ref{app:matched-comparisons}).

\subsection{Signal-Clipping Control}
\label{app:stability-clipping}
The clipping result in Section~\ref{sec:stability} shows that limiting OPD
advantage magnitudes can suppress late instability without teacher takeover.
Here we specify the clamp and compare tight ($b=1$) and loose ($b=5$) bounds.

At each position $t$, the
advantage $A^{\mathrm{OPD}}_{t,v}$ in
Equation~\ref{eq:opd-topk-advantage} weights the teacher--student
log-probability gap by the student's renormalized top-$K$ probability of candidate
$v\in\mathcal V_t$. We clip each of these candidate-level advantages to
$[-b,b]$:
\begin{equation}
  A^{(b)}_{t,v}
  = \min\!\left\{b,\max\!\left\{-b,A^{\mathrm{OPD}}_{t,v}\right\}\right\},
  \qquad b\in\{1,5\}.
  \label{eq:app-signal-clipping}
\end{equation}
Values inside the interval remain unchanged; larger magnitudes are capped
at $b$ while retaining their sign. We substitute $A^{(b)}_{t,v}$ into the
OPD surrogate in Equation~\ref{eq:opd-surrogate}, keeping rollout generation,
policy-ratio clipping, and global gradient clipping unchanged. The clamp
therefore acts on the probability-weighted advantages before loss aggregation.

Table~\ref{tab:app-clipping} shows that the tight $b=1$ clamp suppresses the
late gradient spikes at both budgets, whereas $b=5$ leaves the peak and
the number of high-gradient updates nearly unchanged. At update 279, the
tight clamp changes held-out accuracy by $+0.55$ and $-1.08$ points relative
to the unclipped OPD reproduction at 15K and 18K, respectively.

\begin{table}[t]
  \centering
  \caption{Clipped-OPD control at update 279. Accuracy is held-out average@16
  (\%); gradient statistics exclude the first 50 updates and are measured
  before global gradient clipping.}
  \label{tab:app-clipping}
  \small
  \setlength{\tabcolsep}{5pt}
\begin{tabular}{@{}llrrr@{}}
\toprule
Budget & OPD signal & Peak gradient & Updates $>2$ & Accuracy \\
\midrule
15K & Unclipped reproduction & 14.33 & 22 & 54.50 \\
15K & Clamp $[-5,5]$ & 12.48 & 21 & 55.26 \\
15K & Clamp $[-1,1]$ & \best{0.95} & \best{0} & 55.05 \\
\midrule
18K & Unclipped reproduction & 12.42 & 21 & 54.62 \\
18K & Clamp $[-5,5]$ & 11.28 & 21 & 53.72 \\
18K & Clamp $[-1,1]$ & \best{0.96} & \best{0} & 53.54 \\
\bottomrule
\end{tabular}

\end{table}

\subsection{Intervention Dynamics and Exploration-Floor Ablation}
\label{app:stability-controller}

For the length sweep in Section~\ref{sec:stability}, we additionally examine
intervention frequency, trigger timing in OPD shadow runs, and the effect of
removing the random exploration floor.

\paragraph{Intervention dynamics.}
The three length-sweep arms share $\tau=0.7$ and a $5\%$ random exploration
floor, but the divergence branch activates differently across budgets. At 7K,
the realized takeover rate remains between $2\%$ and $8\%$, making this run a
no-harm control for sparse intervention. At 15K, takeover starts at $35\%$,
recedes to the floor by update 100, and reactivates around the late
instability. At 18K, takeover starts at $91\%$ and remains divergence-driven
for most of training before responding again to the late discrepancy increase.
The 15K and 18K runs therefore test adaptive triggering, while 7K establishes
the stable control.

\paragraph{Trigger timing.}
To examine when the trigger responds, OPD shadow runs evaluate the DTDD
trigger on student rollouts without applying takeover or changing the
training objective. At 18K, the would-trigger rate stays at $8$--$10\%$
through update 198, rises to $25\%$ at update 201, and the gradient spike
begins at update 204. Policy entropy leaves its preceding range at update
192. In this run, divergence reacts near the optimization event rather than
providing an earlier warning than entropy.

\paragraph{Removing the exploration floor.}
The length controller includes a $5\%$ random-intervention floor. We fork the
18K run at update 170 and remove this branch. Across six evaluated checkpoints,
the matched held-out accuracy changes by only $-0.05$ point. The random floor
therefore has negligible effect on student-only accuracy in this regime.

\section{Additional Base-Initialization Results}
\label{app:base-results}

\subsection{Learning Curves and Matched-Checkpoint Results}
\label{app:base-dose-response}
\label{app:base-openthoughts}

This section extends Figure~\ref{fig:efficiency} and
Table~\ref{tab:main-results} with per-controller and per-benchmark results
at shared checkpoints, learning-curve summaries, and updates needed to reach
additional accuracy targets.

\paragraph{DAPO-Math matched checkpoints.}
Table~\ref{tab:app-dose} reports per-controller results at updates 120 and
240, including the 45\% restart arm not plotted in
Figure~\ref{fig:efficiency}. The individual endpoints extend the accuracy range summarized in
Section~\ref{sec:efficiency}.

\begin{table*}[!htbp]
  \centering
  \caption{Base-initialization dose comparison on DAPO-Math-17K
  (average@16, \%). Updates 120 and 240 provide common early and late
  checkpoints for all reported arms.}
  \label{tab:app-dose}
  \small
  \setlength{\tabcolsep}{3pt}
\begin{tabular}{@{}llp{0.40\textwidth}rr@{}}
\toprule
Method & Window $C$ & Controller & Update 120 & Update 240 \\
\midrule
OPD & -- & -- & 14.65 & 22.29 \\
\rowcolor{dtddrow}
\dtdd{} rewind & 256 & fixed $\tau=0.2$, length-gated (initial takeover $\approx25\%$) & 14.81 & 21.58 \\
\rowcolor{dtddrow}
\dtdd{} restart & -- & target rate $45\%$ & 20.49 & 21.67 \\
\rowcolor{dtddrow}
\dtdd{} rewind & 64 & fixed $\tau=0.475$ (initial takeover $\approx90\%$) & \best{21.32} & 21.04 \\
\rowcolor{dtddrow}
\dtdd{} rewind & 128 & fixed $\tau=0.362$ (initial takeover $\approx90\%$) & 20.90 & \best{22.40} \\
\rowcolor{dtddrow}
\dtdd{} rewind & 256 & fixed $\tau=0.296$ (initial takeover $\approx90\%$) & 21.23 & 21.39 \\
\bottomrule
\end{tabular}

\end{table*}

For the high-initial-coverage rewind run with $C=128$,
the mean held-out accuracy gain over OPD is $4.18$ percentage points across
12 shared checkpoints through update 240, compared with $0.56$ over the
final five and $0.10$ at the last checkpoint. OPD's maximum-minus-minimum
accuracy over the final five checkpoints is $5.40$ points
(reporting definitions in Appendix~\ref{app:matched-comparisons}).
The larger all-shared gain and smaller late-window gain reflect faster
early learning rather than a higher final plateau.

\paragraph{Per-benchmark results and OpenThoughts3 controls.}
Table~\ref{tab:app-math-full} complements the best-observed OpenThoughts3
results in Table~\ref{tab:main-results} with per-benchmark scores at the
shared update 200. It also adds the per-benchmark DAPO-Math-17K results at
update 240 for selected rewind and restart controls from
Section~\ref{sec:efficiency}.

Table~\ref{tab:app-openthoughts-reporting} extends the checkpoint comparison
in Table~\ref{tab:main-results} to the final five and all shared
OpenThoughts3 checkpoints. All three controller gains remain positive under
these summaries as well as at the last shared checkpoint, update 200.
The final-five gains range from $1.47$ to $1.68$ points, supporting a
sustained late-training improvement beyond the selected best checkpoints.

\begin{table}[!htbp]
  \centering
  \caption{OpenThoughts3 controller gains over OPD through update 200.
  Accuracy differences and OPD ranges are in percentage points; summaries are defined in
  Appendix~\ref{app:matched-comparisons}.}
  \label{tab:app-openthoughts-reporting}
  \footnotesize
  \setlength{\tabcolsep}{5pt}
\begin{tabular}{@{}lrrrrr@{}}
\toprule
\dtdd{} controller & $N$ & Last $\Delta$ & Last-five $\Delta$ & All-shared $\Delta$ & OPD late range \\
\midrule
Fixed $\tau=0.87$ & 10 & $+1.67$ & $+1.68$ & $+1.49$ & 1.46 \\
Rewind, target $90\%\!\rightarrow\!5\%$ & 10 & $+1.51$ & $+1.47$ & $+1.03$ & 1.46 \\
Restart, target $90\%\!\rightarrow\!5\%$ & 10 & $+2.90$ & $+1.53$ & $+1.21$ & 1.46 \\
\bottomrule
\end{tabular}

\end{table}

\begin{table*}[t]
  \centering
  \caption{Per-benchmark base-initialization results (average@16, \%) at
  shared update 240 for DAPO-Math-17K and 200 for OpenThoughts3.}
  \label{tab:app-math-full}
  \footnotesize
  \begin{tabular}{lrrrr}
\toprule
Method & AIME24 & AIME25 & AMC23 & Average \\
\midrule
\addlinespace
\multicolumn{5}{l}{\emph{Qwen3-1.7B-Base $\leftarrow$ Qwen3-4B-Base-GRPO, DAPO-Math-17K, 7,168 tokens}}\\
\textit{Teacher: Qwen3-4B-Base-GRPO} & \textit{24.17} & \textit{18.12} & \textit{64.22} & \textit{35.50} \\
\opd{} & 11.25 & \best{9.38} & 46.25 & 22.29 \\
\dtdd{} restart, target rate $45\%$ & 11.25 & 7.50 & 46.25 & 21.67 \\
\dtdd{} rewind, initial takeover $\approx90\%$, $C=128$ & 11.67 & 8.33 & \best{47.19} & \best{22.40} \\
\dtdd{} fixed $\tau{=}0.2$, initial takeover $\approx25\%$ & \best{11.88} & 7.71 & 45.16 & 21.58 \\
\addlinespace
\multicolumn{5}{l}{\emph{Qwen3-1.7B-Base $\leftarrow$ Qwen3-4B-Base (non-thinking), OpenThoughts3, 7,168 tokens}}\\
\textit{Teacher: Qwen3-4B-Base} & \textit{21.04} & \textit{20.62} & \textit{67.66} & \textit{36.44} \\
\opd{} & 5.62 & \best{5.42} & 27.50 & 12.85 \\
\dtdd{} rewind, target rate $90\%\!\to\!5\%$ & \best{8.96} & 4.58 & 29.53 & 14.36 \\
\dtdd{} restart, target rate $90\%\!\to\!5\%$ & 8.75 & 5.21 & \best{33.28} & \best{15.75} \\
\dtdd{} fixed $\tau{=}0.87$ & 6.88 & \best{5.42} & 31.25 & 14.51 \\
\bottomrule
\end{tabular}

\end{table*}

\paragraph{Accuracy targets and learning speed.}
Table~\ref{tab:app-crossing} extends the $21.3\%$ reference-level comparison
in Figure~\ref{fig:efficiency} to two higher accuracy targets: OPD's
update-240 scores of $22.29\%$ for seed 0 and $22.75\%$ for the two-seed
mean. It reports the first evaluated update reaching each target.
Among the rewind runs, $C=64$ and $C=256$ reach $22.29\%$ at updates 60 and
100, respectively, compared with update 200 for OPD seed 0. The $C=128$ run
reaches this level at update 220. At the $22.75\%$ target, $C=256$ and the
$45\%$ restart arm cross at updates 180 and 220, while $C=64$ and $C=128$
remain below the target through update 240. Thus, the window-size sweep shows
consistent early accuracy gains, while the time to match OPD's endpoint
depends on the controller and target accuracy.

\begin{table}[t]
  \centering
  \caption{First evaluated update reaching each accuracy level on
  DAPO-Math-17K (average@16, \%). The two OPD-derived targets use update 240;
  ``--'' denotes a target not reached by that update. $^{\ast}$After update
  120, OPD seed 1 is evaluated every 40 updates. Crossing times reflect the
  available evaluation checkpoints.}
  \label{tab:app-crossing}
  \small
  \setlength{\tabcolsep}{5pt}
\begin{tabular}{@{}lrrrr@{}}
\toprule
Arm & $\ge 21.3\%$ & $\ge 22.29\%$ & $\ge 22.75\%$ & Best (step) \\
 & (reference) & (OPD seed 0) & (OPD mean, $n=2$) & \\
\midrule
OPD, seed 0 & 180 & 200 & -- & 22.41 (220) \\
OPD, seed 1$^{\ast}$ & 240 & 240 & 240 & 23.21 (240) \\
\rowcolor{dtddrow}
\dtdd{} rewind, $C=64$ & 60 & 60 & -- & 22.43 (60) \\
\rowcolor{dtddrow}
\dtdd{} rewind, $C=128$ & 80 & 220 & -- & 22.57 (220) \\
\rowcolor{dtddrow}
\dtdd{} rewind, $C=256$ & 70 & 100 & 180 & 22.88 (180) \\
\rowcolor{dtddrow}
\dtdd{} restart, 45\% & 100 & 100 & 220 & 22.97 (220) \\
SFT + OPD & 20 & 80 & 80 & 22.81 (80) \\
\bottomrule
\end{tabular}

\end{table}

\subsection{Controller Dynamics and Design Ablations}
\label{app:base-controls}
\label{app:base-window-form}

This section supplements Figure~\ref{fig:efficiency} and
Table~\ref{tab:main-base-controls} with realized intervention rates,
controller dynamics, trigger and loss details, and individual-seed results.

\paragraph{Realized intervention rates.}
Realized coverage differs from the initial or target rate used to label each
arm. The length-gated control rises from approximately 25\% to 94\%
rollout-level takeover as rollouts lengthen, whereas the 45\% rate-controlled
restart arm realizes approximately 37\% after eligibility filtering.

\begin{figure*}[t]
  \centering
  \includegraphics[width=0.95\textwidth]{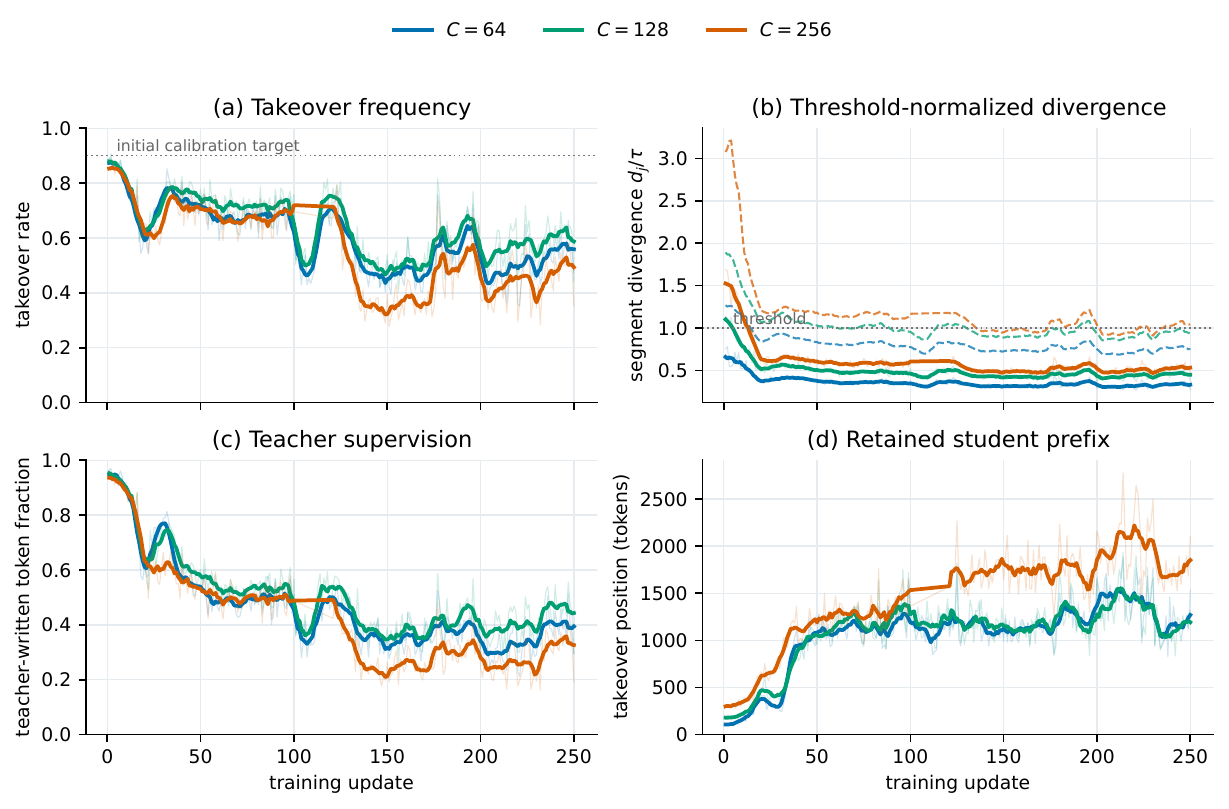}
  \caption{Controller dynamics for the three high-coverage DAPO-Math rewind
  arms. Faint curves show per-update values and solid curves show seven-update
  moving averages. \textbf{(a)} Rollout-level takeover rate; the dotted line is
  the initial calibration target. \textbf{(b)} Segment divergence normalized
  by each arm's fixed threshold (solid: mean; dashed: 90th percentile).
  \textbf{(c)} Fraction of generated training tokens written by the teacher.
  \textbf{(d)} Mean takeover position, conditional on triggering and converted
  from window index to token position.}
  \label{fig:app-dapo-dynamics}
  \vspace{-3mm}
\end{figure*}

\paragraph{Window-size dynamics.}
Figure~\ref{fig:app-dapo-dynamics} complements the accuracy curves in
Figure~\ref{fig:efficiency} with controller measurements for
$C\in\{64,128,256\}$: takeover rate, normalized segment divergence,
teacher-written token fraction, and takeover position. These measurements
separate how often takeover occurs from how much of each triggered rollout
the teacher writes. The normalized divergence curves also show how
student--teacher discrepancy evolves relative to each controller's fixed
threshold.

\paragraph{Trigger and suffix-loss controls.}
\label{app:trigger-loss-controls}
Table~\ref{tab:app-base-controls} expands Table~\ref{tab:main-base-controls}
with evaluations at updates 20 and 40 and the individual OPD and \dtdd{}
seeds underlying its two-seed means. The following analyses add implementation
details, intervention statistics, runtime, and seed-to-seed variation.

The suffix-loss controls in Table~\ref{tab:main-base-controls} also differ
in their supervision support. In the segment-plus-unified-KD row, the
reverse-KL surrogate is evaluated on the
student's top-$K$ support, which can exclude the token written by the teacher.
This arm therefore measures support mismatch as well as the choice between a
unified and role-split objective.
The controllers also differ in realized intervention position and teacher-token
dose (Table~\ref{tab:app-controller-cost}). We therefore interpret
Table~\ref{tab:app-base-controls} as a comparison of complete controller
designs, not an isolated causal estimate of trigger granularity or suffix
objective.

Table~\ref{tab:app-controller-cost} extends the teacher-token column of
Table~\ref{tab:main-base-controls} with trigger frequency, median trigger
position, and runtime. In these later-training measurements,
AdaSwitch~\citep{adaswitch} triggers on every rollout, with a median trigger
at token 14; \dtdd{} triggers on 71\% of rollouts, with a median trigger at
segment 5.

\begin{table*}[t]
  \centering
  \caption{Base-student controller and objective controls on DAPO-Math-17K
  (held-out average@16, \%). A dash denotes an unscheduled evaluation.}
  \label{tab:app-base-controls}
  \footnotesize
  \setlength{\tabcolsep}{3.2pt}
\begin{tabular}{@{}llrrrrrr@{}}
\toprule
Trigger / method & Teacher-suffix objective & \multicolumn{6}{c}{Optimizer step} \\
\cmidrule(l){3-8}
 & & 20 & 40 & 60 & 80 & 100 & 120 \\
\midrule
OPD (seed 0) & -- & 8.58 & 12.67 & 12.36 & 14.01 & 13.47 & 14.65 \\
\rowcolor{dtddrow}
Segment, \dtdd{} (seed 0) & BC & 12.83 & 18.82 & 20.49 & 21.60 & 20.90 & 20.90 \\
OPD (seed 1) & -- & 9.69 & 10.36 & 11.91 & 13.99 & 12.83 & 14.43 \\
\rowcolor{dtddrow}
Segment, \dtdd{} (seed 1) & BC & 12.47 & 19.65 & 20.16 & 21.34 & 21.25 & 20.76 \\
Token spike & Unified KD & 8.84 & 9.53 & 12.47 & 12.14 & 12.81 & 13.51 \\
Token spike & BC & 8.61 & 10.31 & 13.19 & 13.07 & 13.26 & 16.79 \\
Segment & Unified KD & 6.89 & 4.77 & 5.78 & 7.80 & 8.04 & 10.23 \\
AdaSwitch ($K=3$) & Unified KD & 16.35 & 18.94 & 18.16 & 19.57 & 18.68 & 19.95 \\
\bottomrule
\end{tabular}

\end{table*}

\begin{table*}[t]
  \centering
  \caption{Later-training controller and runtime measurements for the
  base-student controls. Trigger position is reported in each controller's
  native granularity. Runtime follows each completed job's allocation; GPU
  models are shown where recorded, and times are not hardware-normalized.}
  \label{tab:app-controller-cost}
  \footnotesize
  \setlength{\tabcolsep}{5pt}
\begin{tabular}{@{}p{0.23\textwidth}p{0.11\textwidth}p{0.09\textwidth}p{0.18\textwidth}p{0.25\textwidth}@{}}
\toprule
Controller / objective & Median trigger & Trigger rate & Teacher-written tokens/update & Time/update \\
\midrule
\rowcolor{dtddrow}
\raggedright \dtdd{} (segment + BC) & segment 5 & 71\% & 0.52M & 15.1 min (4$\times$A100); 19.5 min (4$\times$L40S) \\
\raggedright Token spike + unified KD & token 91 & 79\% & 0.40M & 10.2 min \\
\raggedright Token spike + BC & token 88 & 82\% & 0.43M & 13.0 min \\
\raggedright Segment + unified KD & segment 1 & 80\% & 0.67M & 14.2 min \\
\raggedright AdaSwitch ($K=3$) & token 14 & 100\% & 0.86M & 31.3 min \\
OPD & -- & -- & 0 & 5.4 min (2$\times$A100) \\
\bottomrule
\end{tabular}

\end{table*}

\paragraph{Training-seed replication.}
\label{app:base-seeds}

Table~\ref{tab:app-seed-stats} adds sample standard deviations ($n-1$ in the
denominator) to the two-seed means in Table~\ref{tab:main-base-controls},
and includes the earlier evaluations at updates 20 and 40.
From update 60 onward the seed-to-seed
standard deviation is at most $0.45$ points for both methods, against a gap of
$6.3$--$8.2$ points between them.

\begin{table}[t]
  \centering
  \caption{Two-seed mean$_{\pm\mathrm{SD}}$ on DAPO-Math-17K (average@16, \%).}
  \label{tab:app-seed-stats}
  \small
  \setlength{\tabcolsep}{4pt}
\begin{tabular}{@{}lrrrrrr@{}}
\toprule
Method & \multicolumn{6}{c}{Optimizer step} \\
\cmidrule(l){2-7}
 & 20 & 40 & 60 & 80 & 100 & 120 \\
\midrule
OPD & $9.13_{\pm0.79}$ & $11.52_{\pm1.63}$ & $12.14_{\pm0.32}$ & $14.00_{\pm0.01}$ & $13.15_{\pm0.45}$ & $14.54_{\pm0.16}$ \\
\rowcolor{dtddrow}
\dtdd{} & $12.65_{\pm0.25}$ & $19.23_{\pm0.59}$ & $20.32_{\pm0.23}$ & $21.47_{\pm0.18}$ & $21.08_{\pm0.25}$ & $20.83_{\pm0.10}$ \\
\bottomrule
\end{tabular}

\end{table}

\subsection{Offline SFT Baseline and Computational Cost}
\label{app:base-cost}

This section extends the SFT comparison in Table~\ref{tab:main-base-controls}
and Section~\ref{sec:efficiency} with earlier and later checkpoints,
stage-level timing, and hardware records.

\paragraph{Learning curves.}
Table~\ref{tab:app-warmstart} extends the comparison at updates 60--120 in
Table~\ref{tab:main-base-controls} to updates 20--240. The direct OPD and
\dtdd{} rows report seed 0, while the offline SFT baseline is the same run
as in the main table. SFT followed by OPD reaches
$21.55\%$ at its first evaluation, update 20, and ranges from $20.6\%$ to
$22.8\%$ through update 240. Most of its observed gain is present by the first
evaluated OPD checkpoint.

\begin{table*}[t]
  \centering
  \caption{Fixed-update comparison of direct OPD, online takeover, and offline
  SFT warm start (held-out average@16, \%). A dash denotes an unscheduled
  checkpoint.}
  \label{tab:app-warmstart}
  \small
  \setlength{\tabcolsep}{3.5pt}
\begin{tabular}{@{}lrrrrrrrrr@{}}
\toprule
Training route & \multicolumn{9}{c}{Optimizer step} \\
\cmidrule(l){2-10}
 & 20 & 40 & 60 & 80 & 100 & 120 & 160 & 200 & 240 \\
\midrule
OPD from base & 8.58 & 12.67 & 12.36 & 14.01 & 13.47 & 14.65 & 17.01 & 22.29 & 22.29 \\
\rowcolor{dtddrow}
\dtdd{} from base & 12.83 & 18.82 & 20.49 & 21.60 & 20.90 & 20.90 & 21.02 & 21.96 & 22.40 \\
SFT $\rightarrow$ OPD & 21.55 & 21.27 & 20.63 & 22.81 & 22.45 & 22.03 & 22.31 & 21.67 & 21.77 \\
\bottomrule
\end{tabular}

\end{table*}

\paragraph{Comparison with direct OPD.}
The per-update timings in Table~\ref{tab:app-controller-cost} complement the
update-based comparison in Figure~\ref{fig:efficiency}. Later-training updates
take 5.4 minutes for OPD on two A100 GPUs, compared with 15.1 minutes for \dtdd{} on
four A100 GPUs and 19.5 minutes on four L40S GPUs. These measurements use
different resource allocations and are reported as observed rather than
hardware-normalized costs.

\paragraph{End-to-end pipeline cost.}
Table~\ref{tab:app-pipeline-cost} decomposes the serial-time comparison in
Section~\ref{sec:efficiency} into data generation, SFT, checkpoint merging,
and online training, and adds accelerator use. The offline route spends
26.9 hours generating teacher data and 11.7 hours on SFT before OPD begins.
Measured accelerator use is approximately 66 GPU-hours for the offline route
and 88 GPU-hours for online \dtdd{}.
Generation, SFT, and checkpoint merging use completed-job elapsed times;
OPD and \dtdd{} use cumulative logged step times through updates 20 and 80,
respectively. These costs exclude queue delays and failed launches. Totals
use unrounded measurements.
These wall-clock values use the hardware assigned to each completed job---A40
and A100 GPUs for the SFT route, and four L40S GPUs for
\dtdd{}---and are not normalized across device types.

\begin{table*}[t]
  \centering
  \caption{Serial cost to comparable observed checkpoints. Training time
  sums logged step times. Stage times are rounded; totals use unrounded
  measurements. Accelerator types differ across routes.}
  \label{tab:app-pipeline-cost}
  \footnotesize
  \setlength{\tabcolsep}{3pt}
\begin{tabular}{@{}p{0.15\textwidth}p{0.25\textwidth}p{0.13\textwidth}p{0.11\textwidth}p{0.08\textwidth}p{0.17\textwidth}@{}}
\toprule
Route & Measured stages & Checkpoint (accuracy) & Serial time & GPU-h & Teacher-written tokens \\
\midrule
\rowcolor{dtddrow}
Online \dtdd{} (seed 1) & online training (22.0 h) & Step 80 (21.34) & 22.0 h & 88 & 43M \\
SFT $\rightarrow$ OPD & teacher data (26.9 h) $+$ SFT (11.7 h) $+$ merge (0.1 h) $+$ OPD (4.0 h) & Step 20 (21.55) & $\approx$42.8 h & $\approx$66 & 95M \\
\bottomrule
\end{tabular}

\end{table*}

\section{Additional Agentic Results}
\label{app:agentic-results}

\subsection{ALFWorld Action-Level Intervention}
\label{app:alfworld-intervention}

Table~\ref{tab:app-agentic-endpoint} extends the ALFWorld comparison in
Table~\ref{tab:main-results} to the final shared update, 400, and adds a
full-model training control alongside LoRA.
At update 400, LoRA improves average success from 70.06\% to 77.59\%, driven
primarily by the seen split. Full-model training raises OPD to 79.50\% and
narrows the \dtdd{} advantage to about 0.4 point
(Table~\ref{tab:app-agentic-endpoint}). Parameter-efficient adaptation
therefore leaves a larger recoverable gap, while full-model OPD closes most of
that gap on its own.

\begin{table}[!htbp]
  \centering
  \caption{ALFWorld success rate (\%) at update 400, the last common
  checkpoint. Full-model training provides a capacity control for LoRA.}
  \label{tab:app-agentic-endpoint}
  \small
  \begin{tabular}{lrrr}
\toprule
Method & Seen & Unseen & Average \\
\midrule
\addlinespace
\multicolumn{4}{l}{\emph{Qwen2.5-3B-Instruct $\leftarrow$ GiGPO-7B, LoRA}}\\
\opd{} & 70.71 & 69.40 & 70.06 \\
\dtdd{} & \best{84.29} & \best{70.90} & \best{77.59} \\
\addlinespace
\multicolumn{4}{l}{\emph{Qwen2.5-3B-Instruct $\leftarrow$ GiGPO-7B, full fine-tuning}}\\
\opd{} & \best{82.14} & 76.87 & 79.50 \\
\dtdd{} & \best{82.14} & \best{77.61} & \best{79.88} \\
\bottomrule
\end{tabular}

\end{table}

Across the last five matched checkpoints, \dtdd{} leads by $7.22$ points on
average. The OPD arm
varies by $10.2$ points over the same window, so repeated runs are needed to
separate the intervention effect from late-training variation.

\begin{figure*}[t]
  \centering
  \includegraphics[width=0.80\textwidth]{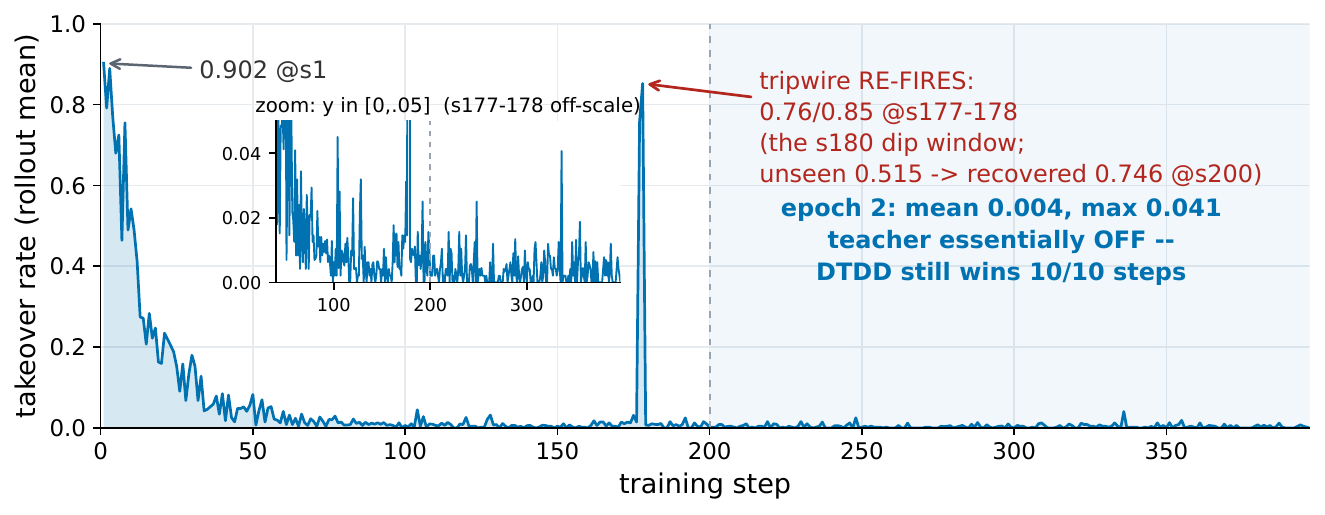}
  \caption{ALFWorld action-level takeover (Qwen2.5-3B-Instruct $\leftarrow$
  GiGPO-7B). The controller withdraws as discrepancy contracts and reactivates
  around the update-180 disruption.}
  \label{fig:agentic-dynamics}
  \vspace{-3mm}
\end{figure*}

\begin{figure*}[t]
  \centering
  \includegraphics[width=0.94\textwidth]{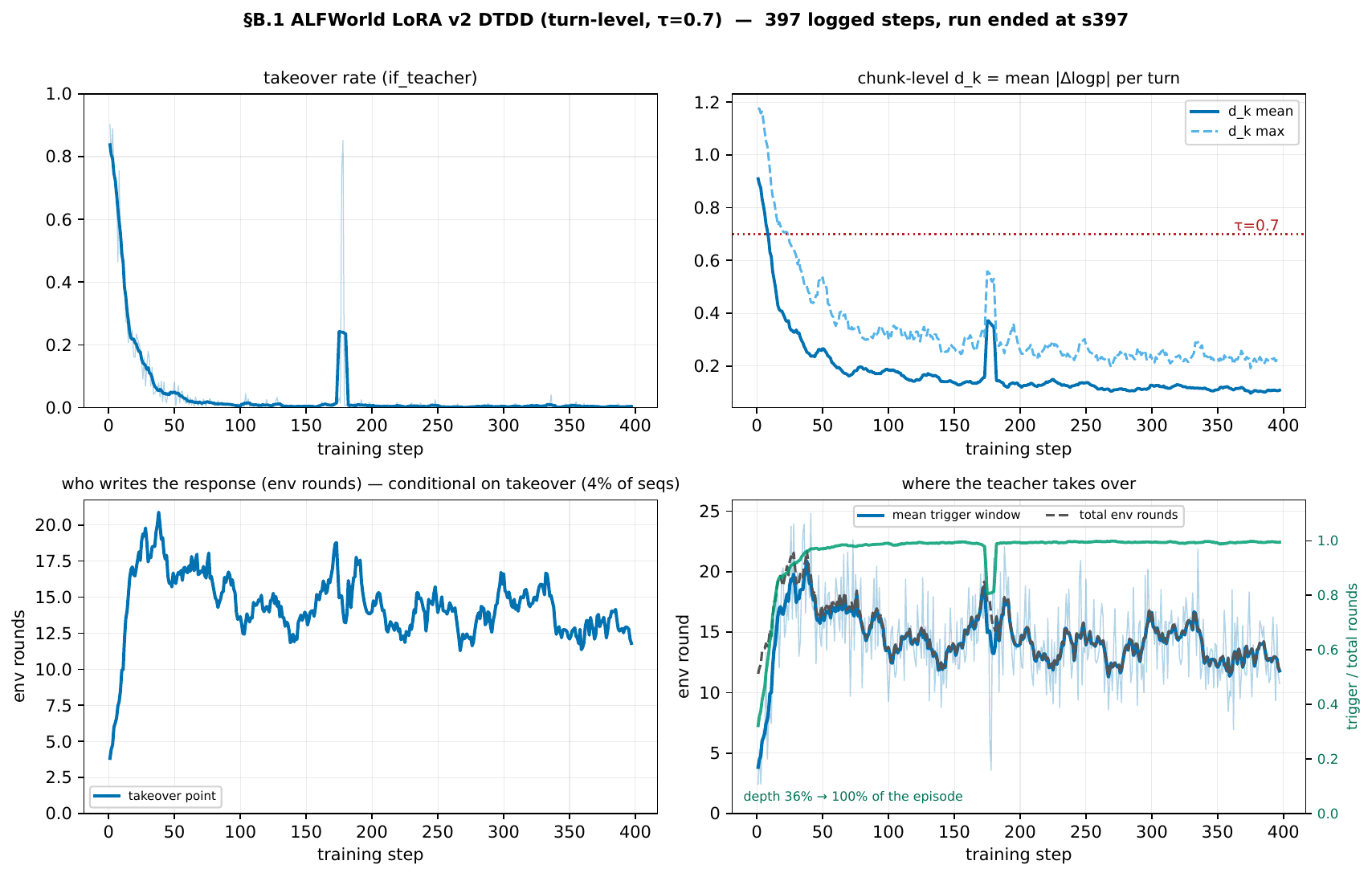}
  \caption{ALFWorld action-level controller dynamics. Top: takeover rate and
  action discrepancy. Bottom: intervention turn and its depth within the
  episode, conditional on takeover. The student is Qwen2.5-3B-Instruct, the
  teacher is GiGPO-7B, and the maximum episode length is 30 turns.}
  \label{fig:app-alfworld-dynamics}
  \vspace{-3mm}
\end{figure*}

Figures~\ref{fig:agentic-dynamics} and~\ref{fig:app-alfworld-dynamics}
provide the controller measurements behind the ALFWorld analysis in
Section~\ref{sec:agentic-results}. The first resolves the low takeover rates
and brief reactivation; the second adds action discrepancy, intervention turn,
and relative episode depth, showing where the remaining interventions occur.

\subsection{Agentic Code Dose}
\label{app:agentic-code-dose}

Table~\ref{tab:app-code-endpoint} complements the best-observed code results
in Table~\ref{tab:main-results} with the final shared update, 130, and a
higher-dose control ($\tau=0.26$) for the same student--teacher pair.
At the shared update-130 endpoint, takeover with $\tau=0.50$ remains $1.03$
points above OPD in aggregate because of gains on the two in-distribution
splits, although both LiveCodeBench scores trail OPD
(Table~\ref{tab:app-code-endpoint}).

The two thresholds deliver $2.8\%$ and approximately $50\%$ teacher-token dose.
Increasing the dose to approximately $50\%$ reverses the gain over OPD,
reducing aggregate accuracy from $33.93\%$ to $28.13\%$ at update 130. The
teacher itself reaches
60.63\% on the same aggregate evaluation, so teacher quality leaves ample
headroom. The contrasting outcome points to excessive replacement of a
warm-start student's on-policy interaction.

\begin{table}[t]
  \centering
  \caption{Agentic code execution accuracy (average@8, \%) at update 130.
  Held-out averages the two in-distribution splits; LCB averages LiveCodeBench
  v1 and v6.}
  \label{tab:app-code-endpoint}
  \scriptsize
  \begin{tabular}{lrrrr}
\toprule
Method & Dose & Held-out & LCB & Avg. \\
\midrule
\addlinespace
\multicolumn{5}{l}{\emph{Qwen3-1.7B-SFT $\leftarrow$ DemyAgent-GRPO-4B, code interaction}}\\
Teacher & -- & 63.16 & 58.11 & 60.63 \\
\opd{} & 0\% & 25.60 & \best{42.26} & 33.93 \\
\dtdd{} ($\tau=0.50$) & 2.8\% & \best{31.50} & 38.43 & \best{34.96} \\
\dtdd{} ($\tau=0.26$) & $\sim$50\% & 19.78 & 36.48 & 28.13 \\
\bottomrule
\end{tabular}

\end{table}

\section{Headroom Analysis and Boundary Cases}
\label{app:boundary-results}

This appendix supplements Figure~\ref{fig:headroom} and
Section~\ref{sec:boundaries} with the per-arm measurements underlying the
headroom plot, detailed boundary experiments, and a reanalysis of published
TCOD results.

\subsection{Headroom Inventory and Reporting Details}
\label{app:headroom-inventory}

Table~\ref{tab:app-headroom-inventory} provides the teacher scores, OPD
endpoints, paired-checkpoint counts, and mean gains for all 23 arms from the
13 settings in Figure~\ref{fig:headroom}. Headroom follows the definition in
Section~\ref{sec:boundaries}; gains average the \dtdd{}--OPD difference
over the matched checkpoints available for each arm. This differs from the
best-observed protocol of Table~\ref{tab:main-results}: the
ALFWorld LoRA arm,
for example, has $19.4\%$ headroom and a $+1.11$-point matched-checkpoint mean
here, against a $+4.83$-point best-observed gap in
Table~\ref{tab:main-results}.
Within each baseline group, the OPD endpoint is
the last checkpoint shared by OPD and all full-length \dtdd{} arms. The
shorter 25\%-target control and the 18K exploration-floor ablation use the same
group headroom but contribute only their available matched checkpoints to the
mean gain. Each included arm has at least three paired evaluations.

The analysis is descriptive: arms that share a setting are not independent,
and the $10\%$ boundary summarizes the observed split rather than defining a
fitted decision threshold.
The split summarizes the observed sign rather than the magnitude. Six
DAPO-Math arms share one headroom value but range from $+0.11$ to $+4.42$
points as dose and intervention form change.

\begin{table*}[t]
  \centering
  \caption{Complete inventory for Figure~\ref{fig:headroom}. Scores and
  headroom $h$ are percentages; mean $\Delta$ is the \dtdd{}--OPD difference
  in percentage points across $N$ matched checkpoints. Step identifies the
  OPD endpoint used to compute headroom. Math uses average@16, ALFWorld and
  WebShop use success rate, and code uses LiveCodeBench v1 average@8.}
  \label{tab:app-headroom-inventory}
  \footnotesize
\setlength{\tabcolsep}{4pt}
\begin{tabular*}{\textwidth}{@{\extracolsep{\fill}}lrrrrrr@{}}
\toprule
Arm & Teacher & OPD endpoint & Step & $h$ (\%) & $N$ & Mean $\Delta$ \\
\midrule
\multicolumn{7}{@{}l}{\textit{DAPO-Math: R1-Distill-1.5B student / JustRL-1.5B teacher}} \\
7K rewind                  & 58.45 & 54.44 & 260 &  6.86 & 10 & $-0.04$ \\
15K rewind                 & 58.45 & 54.15 & 279 &  7.37 & 10 & $-0.62$ \\
18K rewind                 & 58.45 & 55.38 & 279 &  5.26 & 14 & $-1.20$ \\
18K, no exploration floor  & 58.45 & 55.38 & 279 &  5.26 &  6 & $-0.53$ \\
Hard-gap, fixed threshold  & 58.45 & 53.12 & 150 &  9.12 &  7 & $-0.47$ \\
Hard-gap, scheduled restart& 58.45 & 53.12 & 150 &  9.12 &  8 & $-1.07$ \\
\midrule
\multicolumn{7}{@{}l}{\textit{DAPO-Math: Qwen3-1.7B-Base student / Qwen3-4B-Base-GRPO teacher}} \\
25\%-target control        & 35.50 & 22.29 & 240 & 37.21 &  6 & $+0.11$ \\
Fixed $\tau=0.2$, length-gated & 35.50 & 22.29 & 240 & 37.21 & 12 & $+0.15$ \\
45\%-target restart        & 35.50 & 22.29 & 240 & 37.21 & 12 & $+3.50$ \\
90\% initial, $C=64$       & 35.50 & 22.29 & 240 & 37.21 & 12 & $+4.19$ \\
90\% initial, $C=128$      & 35.50 & 22.29 & 240 & 37.21 & 12 & $+4.18$ \\
90\% initial, $C=256$      & 35.50 & 22.29 & 240 & 37.21 & 12 & $+4.42$ \\
\midrule
\multicolumn{7}{@{}l}{\textit{DAPO-Math: Qwen3-1.7B-SFT student / Qwen3-4B-Base-GRPO teacher}} \\
SFT-initialized student    & 35.50 & 20.75 & 279 & 41.56 &  4 & $-0.36$ \\
\midrule
\multicolumn{7}{@{}l}{\textit{OpenThoughts3: Qwen3-1.7B-Base student / Qwen3-4B teacher}} \\
Scheduled rewind          & 36.44 & 12.85 & 200 & 64.75 & 10 & $+1.03$ \\
Scheduled restart         & 36.44 & 12.85 & 200 & 64.75 & 10 & $+1.21$ \\
Fixed $\tau=0.87$         & 36.44 & 12.85 & 200 & 64.75 & 10 & $+1.49$ \\
\midrule
\multicolumn{7}{@{}l}{\textit{DAPO-Math: R1-Distill-1.5B student / R1-Distill-7B teacher}} \\
Stable-OPD setting         & 59.31 & 42.31 & 279 & 28.66 & 14 & $-0.08$ \\
\midrule
\multicolumn{7}{@{}l}{\textit{ALFWorld: Qwen2.5-3B-Instruct student / GiGPO-7B teacher}} \\
LoRA                      & 86.89 & 70.06 & 400 & 19.37 & 20 & $+1.11$ \\
Full fine-tuning           & 86.89 & 79.50 & 400 &  8.50 & 20 & $+5.25$ \\
\midrule
\multicolumn{7}{@{}l}{\textit{ALFWorld: Qwen2.5-0.5B-Instruct student / GiGPO-7B teacher}} \\
Wide teacher--student gap  & 86.89 &  3.65 & 120 & 95.80 &  6 & $+0.17$ \\
\midrule
\multicolumn{7}{@{}l}{\textit{WebShop: Qwen2.5-3B-Instruct student / GiGPO-7B teacher}} \\
Rebuilt 120K catalog       & 18.69 & 21.21 & 180 & $-13.51$ & 9 & $-7.31$ \\
\midrule
\multicolumn{7}{@{}l}{\textit{Agentic code: Qwen3-1.7B-SFT student / DemyAgent-GRPO-4B teacher}} \\
Sparse, $\tau=0.50$       & 68.78 & 51.37 & 130 & 25.31 & 11 & $+2.93$ \\
High-dose, $\tau=0.26$    & 68.78 & 51.37 & 130 & 25.31 & 12 & $-7.48$ \\
\bottomrule
\end{tabular*}

\end{table*}

Four arms fall outside the common split. ALFWorld full fine-tuning averages a
$5.25$-point gain across 20 matched checkpoints but finishes only about $0.4$
point above OPD, indicating transient acceleration. Both axes for the code
points use LiveCodeBench v1. The $\tau=0.26$ arm loses $7.48$ points when its
LiveCodeBench-v1 gap is averaged over 12 matched checkpoints. The $-5.80$-point
result in Table~\ref{tab:app-code-endpoint} instead compares OPD and \dtdd{} at
their shared update-130 endpoint after averaging HO-200 full, HO-200 subset,
LiveCodeBench v1, and LiveCodeBench v6. The same arm has an $85.4\%$
rollout-level takeover rate (approximately $50\%$ teacher-token dose), showing
that excessive replacement can erase otherwise useful headroom. The
``SFT student'' point is Qwen3-1.7B-SFT $\leftarrow$
Qwen3-4B-Base-GRPO on DAPO-Math-17K; the ``stable-OPD lane'' is
DeepSeek-R1-Distill-Qwen-1.5B $\leftarrow$
DeepSeek-R1-Distill-Qwen-7B on DAPO-Math-17K. Their matched-checkpoint means
differ from OPD by $-0.36$ and $-0.08$ point, respectively.

\subsection{Boundary Cases}
\label{app:headroom-boundaries}

The teacher-validity and student-learnability controls summarized in
Section~\ref{sec:boundaries} are detailed below, including the environment
changes and matched evaluations. Table~\ref{tab:app-boundaries} collects
their settings and quantitative outcomes alongside the dose controls.

\begin{table*}[!htbp]
  \centering
  \caption{Boundary cases and their empirical interpretation. Scores use
  the native evaluation metric of each environment.}
  \label{tab:app-boundaries}
  \footnotesize
  \setlength{\tabcolsep}{4pt}
\begin{tabular}{@{}p{0.17\textwidth}p{0.24\textwidth}p{0.21\textwidth}p{0.14\textwidth}p{0.15\textwidth}@{}}
\toprule
Condition & Setting & Observation & Result & Interpretation \\
\midrule
Little remaining gap
  & 7K/15K/18K distilled math student
  & OPD finishes within 5--9\% of teacher
  & Stability gain; accuracy tie
  & Stability without higher accuracy \\
Valid, learnable gap
  & Base math student
  & OPD learns slowly; teacher continuation is learnable
  & High dose reaches the reference score earlier
  & Faster early learning \\
Warm, healthy student
  & Agentic code, $\rho_T=2.8\%$
  & Small in-distribution gain
  & $+1.03$ points at update 130
  & Sparse intervention can help \\
Excessive dose
  & Agentic code, $\rho_T\approx50\%$
  & Teacher behavior replaces on-policy interaction
  & $-5.80$ aggregate points at update 130
  & Excessive replacement \\
Invalid teacher
  & Rebuilt WebShop
  & Teacher 18.69\% $<$ OPD 21.21\%
  & \dtdd{} 11.62\% at update 180
  & Teacher mismatch \\
Limited learnability
  & ALFWorld, 0.5B $\leftarrow$ 7B
  & Matched mean: OPD $0.79\%$, \dtdd{} $0.96\%$
  & $+0.17$ points; both remain near zero
  & Limited student transfer \\
\bottomrule
\end{tabular}

\end{table*}

\paragraph{Teacher validity.}
The rebuilt WebShop environment uses a 120K-product catalog that retains all
goal-bearing products and adds random distractors. It therefore differs from
the catalog and observation format on which the WebShop-specific GiGPO-7B
teacher was trained. On this environment, that teacher reaches $18.69\%$
success. At the last shared checkpoint, update 180, OPD reaches $21.21\%$
and \dtdd{} reaches $11.62\%$: takeover routes training toward a teacher that is
weaker on the current environment.

\paragraph{Student learnability.}
The small-student ALFWorld experiment pairs Qwen2.5-0.5B-Instruct with the
ALFWorld-specific GiGPO-7B teacher, creating a roughly $14\times$ parameter
gap. Across the six matched evaluations from updates 20 through 120, OPD and
\dtdd{} average $0.79\%$ and $0.96\%$ success, respectively. At update 120,
the scores are $3.65\%$ and $1.80\%$; the additional \dtdd{} evaluation at
update 140 reaches $4.71\%$. The teacher supplies valid behavior, but takeover
does not produce reliable student-only success under this training
configuration.

\paragraph{Intervention dose.}
The dose control holds the code student, teacher, and task
fixed while varying the takeover threshold. Appendix~\ref{app:agentic-code-dose}
reports its realized teacher-token doses and per-split endpoint results.

\subsection{Reanalysis of TCOD Results}
\label{app:tcod-reanalysis}

To examine the pattern in Figure~\ref{fig:headroom} beyond our own runs,
we apply the same headroom calculation to the published main TCOD
results \citep{tcod}. This separates its two student scales. The $1.7$B OPD
baselines retain over $99\%$
headroom, and all 18 takeover cells improve by $9.2$--$24.7$ points. The $4$B
baselines finish within $6.9$--$13.4\%$ of their teachers; gains averaged across
the six curriculum configurations within each benchmark fall to $0.3$--$1.5$
points, with four negative cells. These published results show a
similar qualitative pattern to Figure~\ref{fig:headroom}. This comparison
uses reported scores rather than new TCOD runs.

\end{document}